\documentclass{article}

\PassOptionsToPackage{numbers,sort&compress}{natbib}
\usepackage[preprint]{neurips_2026}

\usepackage[utf8]{inputenc} 
\usepackage[T1]{fontenc}    
\usepackage{hyperref}       
\usepackage{url}            
\usepackage{booktabs}       
\usepackage{amsfonts}       
\usepackage{nicefrac}       
\usepackage{microtype}      
\usepackage{xcolor}         
\input{sections/preamble}

\title{
Past as State, Present as Attention: \\
Persistent-State Blockwise Flow Matching for \\
Long-Horizon Co-Speech Motion Generation
}

\author{%
  Yi Liu$^{1,3}$ \quad Xiangyue Zhang$^2$ \quad
  Jianfang Li$^3$\thanks{Corresponding author.} \quad
  Jia Ma$^1$ \quad Yisheng He$^3$ \quad Jianqiang Ren$^3$ \\
  $^1$ Tongji University \quad
  $^2$ The University of Tokyo \quad
  $^3$ Alibaba Group \\ [0.3em]
  {\small
    \texttt{\{liuyi61,2432018\}@tongji.edu.cn} \quad
    \texttt{x-zhang@mi.t.u-tokyo.ac.jp}
  } \\ 
  {\small
    \texttt{\{wuhui.ljf,heyisheng.hys,jianqiang.rjq\}@alibaba-inc.com}
  } \\ [0.3em]
  {\small
    \added{Project website: \url{https://ly403.github.io/PASPA/}}
  }%
}

\begin{document}

\maketitle
\begin{figure}[h] 
    \centering
    \vspace{-2mm}
    \includegraphics[width=\linewidth]{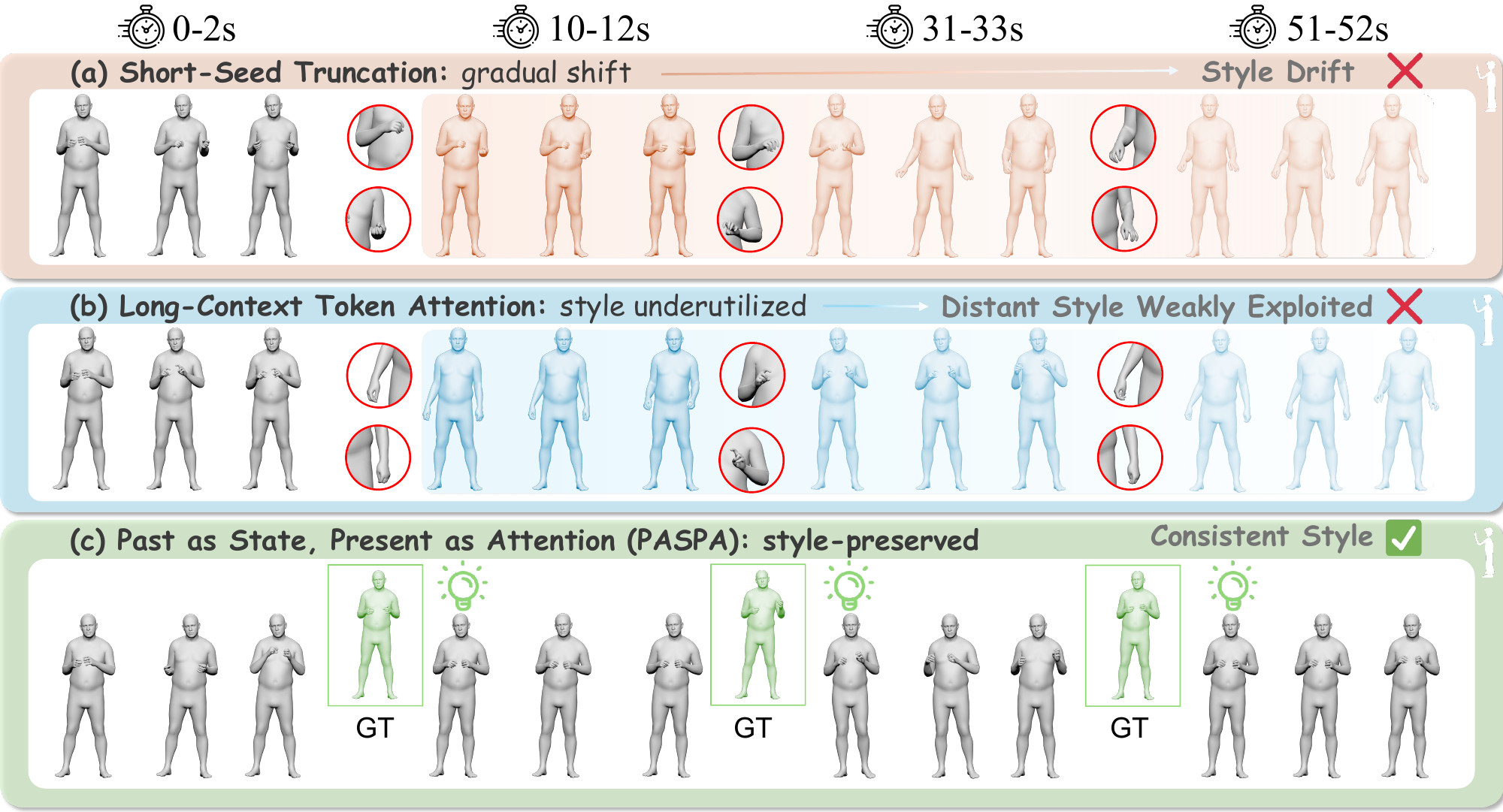}
    \caption{
        Long-horizon co-speech motion generation (52\,s) from the same speech and speaker, with the GT style keeping both hands raised near the chest.
        (a) Short-Seed Truncation matches the style initially but monotonically drifts: hands gradually drop. 
        (b) Long-Context Token Attention retains full history but struggles to maintain a consistent style, matching or deviating regardless of temporal distance. 
        (c) PASPA (ours) preserves the raised-hands style throughout, closely matching the GT.
        \label{fig:teaser}
    }
    \vspace{-2mm}
\end{figure}
\begin{abstract}
Co-speech motion generation remains challenging over long horizons due to the need for temporal coherence and consistent style.
Existing methods typically generate long motions segment by segment, either conditioning on a short seed of recent frames, which gradually drifts, or retaining ever-longer explicit histories, which are costly to maintain and hard to learn from limited data. 
We trace these limitations to a key asymmetry between past and present:
past motion carries slowly varying style that should be compactly summarized, whereas the current segment exhibits fast-changing local dynamics that require fine-grained modeling.
This asymmetry motivates a simple \emph{\textbf{P}ast-\textbf{A}s-\textbf{S}tate, \textbf{P}resent-as-\textbf{A}ttention} principle for long-horizon co-speech motion generation.
Following this principle, we propose \textbf{PASPA}, which summarizes past history with an inter-block persistent state and models the current segment with intra-block bidirectional self-attention.
%
We instantiate PASPA within a blockwise flow matching framework, with multi-block supervision for effective training over variable histories, prefill-decode state caching for efficient rollout, and hybrid classifier-free guidance for enhanced history and speech conditioning.
Extensive public-benchmark experiments show that PASPA achieves state-of-the-art FGD performance and preserves motion style across long-sequence generation. 
\end{abstract}
\section{Introduction}
\label{sec:introduction}
Humans naturally accompany speech with expressive body movements that convey emphasis, emotion, and communicative intent beyond words alone~\cite{mcneill1992hand}.
Co-speech motion generation aims to synthesize realistic full-body motions that are temporally aligned and semantically consistent with speech~\cite{nyatsanga2023comprehensive}.
%
\added{It is crucial to applications like digital humans~\cite{zhu2023human}, virtual assistants~\cite{yoon2020speech}, and animated characters~\cite{habibie2021learning}.}

Early co-speech motion generation often relies on direct speech-to-motion mapping~\cite{yoon2019robots}, but the inherent one-to-many nature of speech-to-motion correspondence has made generative modeling increasingly important~\cite{li2021audio2gestures}.
%
Accordingly, diffusion models~\cite{ho2020denoising} and flow matching~\cite{liu2023flow,lipman2023flow}, the prevailing generative paradigms, have been widely adopted in this field.\footnote{We use diffusion and flow matching interchangeably in this paper due to their mathematical equivalence~\cite{albergo2025stochastic,lipman2024flow}.}
While they improve short-horizon realism and controllability~\cite{zhang2026mitigating,liu2025gesturelsm}, long-horizon co-speech motion generation remains challenging, as it requires preserving temporal coherence~\cite{vu2026streaming} and speaker style~\cite{ahuja2020style} (see \cref{fig:teaser}).

\begin{figure}[t] 
    \centering
    \includegraphics[width=\linewidth]{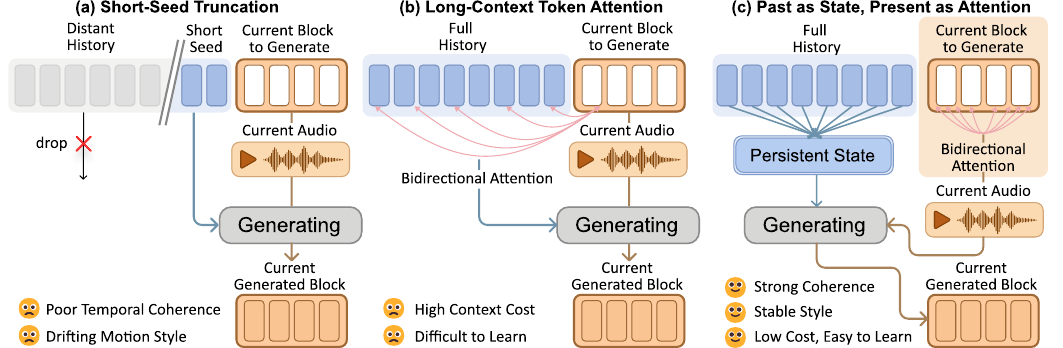}
    \caption{
        %
        %
        History propagation in co-speech motion generation.
        (a) Short-Seed Truncation keeps only recent frames.
        (b) Long-Context Token Attention retains history as tokens for global attention.
        (c) PASPA compresses history into a persistent state and applies full attention within the current block.
        \label{fig:compare}
    }
\end{figure}

For long-horizon generation, most methods proceed segment by segment, making inter-block history propagation the central challenge.
A common strategy is to use only a seed (\ie, the previous segment's last few frames; \cref{fig:compare}a)~\cite{liu2024emage,zhang2025semtalk,zhang2026mitigating} as condition.
%
With such limited history, temporal coherence and style gradually drift over time (see~\cref{fig:teaser}a)~\cite{vu2026streaming}.
Recent streaming methods exploit longer histories~\cite{vu2026streaming,xiao2025motionstreamer,cai2025flooddiffusion}, but retain contexts like frames or tokens (see \cref{fig:compare}b). 
%
Maintaining such explicit histories incurs growing memory and compute costs as horizons extend, especially with attention-based backbones~\cite{vaswani2017attention}.
%
%
More critically, learning to exploit ever longer explicit histories is difficult given the limited data in co-speech motion generation~\cite{liu2024emage}, often leading to unstable style preservation (see \cref{fig:teaser}b).
%
These observations raise a fundamental question: 
\emph{is there a more effective way to represent and propagate the past than explicitly attending to raw historical sequences?}

We answer this affirmatively: the past and present play distinct roles in motion generation and should be represented accordingly.
%
Past history primarily carries slowly varying stylistic traits (\eg, gesture amplitude and spatial preference)~\cite{ahuja2023continual,ahuja2020style}, which should be compactly summarized rather than stored verbatim.
%
By contrast, local motion dynamics are fast-changing, making full-resolution modeling of the current block more suitable for capturing fine-grained motion details.
%
%
This asymmetry motivates a simple \emph{\textbf{P}ast-\textbf{A}s-\textbf{S}tate, \textbf{P}resent-as-\textbf{A}ttention} principle: the past should be compressed as a persistent state that is carried across segments, while the present should be modeled with full attention for expressive generation (see \cref{fig:compare}c).
%
Guided by this principle, we propose \textbf{PASPA}, which achieves temporally coherent and style-consistent long-horizon co-speech motion generation (see \cref{fig:teaser}c).


In PASPA, past history and the current block are handled by two complementary mechanisms. 
%
For past history, we design a block-level recurrent state mechanism built upon Gated DeltaNet~\cite{yang2024parallelizing,yang2025gated}, which progressively summarizes preceding blocks into a persistent state and provides a selective-retention, temporal-decay inductive bias for slowly evolving motion style.
%
For the current block, we use blockwise self-attention~\cite{vaswani2017attention,qiu2020blockwise} to enable full bidirectional interaction for fine-grained modeling.
%
We integrate them into a hybrid PASPA module, which forms the core component of our model. 

We build PASPA on a semi-autoregressive blockwise flow matching (BFM) framework inspired by recent block diffusion paradigms~\cite{arriola2025block,zhang2025blockvid}, and develop tailored training and inference protocols.
%
For training, we apply distinct noise levels to multiple blocks and supervise their denoising in parallel, yielding dense multi-block supervision and exposing each block to states summarizing variable-length histories.
For inference, PASPA separates state prefilling from iterative decoding and reuses cached historical states for efficient rollout.
To improve generation quality, we introduce a hybrid classifier-free guidance~\cite{ho2022classifier} that augments audio-conditioned guidance with history-drop guidance~\cite{song2025history}, enabling explicit control over historical dependence while preserving speech alignment.

Our main technical contributions are:
%
(1) We identify a key asymmetry in the roles of past and present in co-speech motion generation, and formulate a simple \emph{past-as-state, present-as-attention} principle for long-horizon modeling.
%
(2) Following this principle, we propose \textbf{PASPA}, a persistent-state method built on blockwise flow matching, whose core module couples inter-block persistent state modeling with intra-block attention.
(3) We develop tailored training and inference schemes, including multi-block supervision, prefill-decode decoupling with state caching, and hybrid classifier-free guidance.
\section{Related Work}
\label{sec:related_work}

\paragraph{Co-Speech Motion Generation.}
%
The task synthesizes speech-aligned body movements.
Its one-to-many nature motivates generative paradigms such as diffusion models (DMs)~\cite{ho2020denoising,song2019generative} and VQ-VAE-based generation~\cite{van2017neural} for realism and diversity~\cite{habibie2021learning,ahuja2022low,ahuja2023continual,zhu2023taming, zhang2026streamtalk}.
Further enhancements for semantic controllability include hierarchical designs and language-based guidance~\cite{qi2024emotiongesture,liu2022learning,liang2022seeg,chen2024enabling,zhang2026personagesture}.
%
VQ-VAE-based methods~\cite{yi2023generating,liu2024emage,liu2024towards,zhang2025semtalk,zhang2025echomask} generate motions through learned codebooks, but often suffer from limited diversity due to discrete code constraints.
DM-based methods~\cite{chen2024diffsheg,ao2023gesturediffuclip,cheng2025hologest,mughal2025retrieving,zhang2026mitigating} better support diverse motion synthesis and have become a dominant paradigm in this field.

\paragraph{Diffusion Models and Flow Matching.}
DMs~\cite{ho2020denoising} and Flow Matching (FM)~\cite{lipman2023flow,liu2023flow} have achieved strong results in image generation~\cite{croitoru2023diffusion,dhariwal2021diffusion}.
Advances include novel architectures~\cite{peebles2023scalable,esser2024scaling,xie2024sana}, conditional guidance~\cite{ho2022classifier,kynkaanniemi2024applying,karras2024guiding}, and accelerated sampling~\cite{song2021denoising,liu2023flow}.
They also extend to sequential domains (\eg, video) via autoregressive (AR) factorization for variable-length rollout~\cite{arriola2025block,chen2024diffusion,ruhe2024rolling}.
Representative approaches include Block Diffusion~\cite{arriola2025block,zhang2025blockvid}, which combines intra-block denoising with inter-block autoregression, and Diffusion Forcing~\cite{chen2024diffusion}, which assigns independent per-token noise levels to bridge next-token prediction~\cite{radford2018improving} and full-sequence diffusion.
Related advances mitigate exposure bias~\cite{huang2025self}, enable real-time generation~\cite{liu2025rolling}, and support flexible history conditioning~\cite{song2025history}.
This AR diffusion paradigm has also been adapted to motion generation~\cite{han2024amd,zhao2025dartcontrol,cai2025flooddiffusion,xiao2025motionstreamer}. 

\paragraph{Long-Context Sequence Modeling.}
Attention~\cite{vaswani2017attention} dominates long-context modeling but scales quadratically with sequence length~\cite{tay2022efficient} and lacks inductive biases for data-limited regimes~\cite{dosovitskiy2021an}.
%
Alternatives like state space models~\cite{gu2023mamba,dao2024transformers} and linear attention~\cite{katharopoulos2020transformers,yang2024parallelizing,yang2024gated} replace pairwise attention with linear recurrent state updates, improving efficiency and introducing stronger temporal inductive bias.
%
Specifically, Gated DeltaNet (GDN)~\cite{yang2025gated} is highly relevant to us, as its gating and delta-rule updates yield a temporal-decay bias that we  leverage to model continuous motion dynamics. 
%
Such long-context designs have also been adopted in motion generation (\eg, Motion Mamba~\cite{zhang2024motion}).

\section{Methodology}
\label{sec:methodology}
\subsection{Preliminary}
\label{sec:preliminary}

\paragraph{Co-Speech Motion Generation.}
This task synthesizes motion 
$\bfx \in \mathbb{R}^{T \times C}$ and facial expression $\bfe \in \mathbb{R}^{T \times C_e}$ from a speech waveform $\bfa \in \mathbb{R}^{T_a}$.
%
Following the SMPL-X~\cite{loper2015smpl,pavlakos2019expressive} parameterization, the motion sequence $\bfx$ consists of $J=55$ joint rotations and a 3D global translation.
The joint rotations are converted from axis-angle to the 6D representation (rot6d)~\cite{zhou2019continuity}, yielding $C=6J+3=333$.
Following FLAME~\cite{li2017learning}, the expression $\bfe$ is represented by $C_e = 100$ FLAME coefficients.
For long-horizon generation, co-speech motion is commonly generated segment by segment~\cite{zhang2026mitigating,zhang2025semtalk,liu2024emage}.
Accordingly, the motion and expression sequences are partitioned into $N$  blocks of length $K$, denoted by $\bfx = [\bfx^{1}, \ldots, \bfx^{N}]$ and
$\bfe = [\bfe^{1}, \ldots, \bfe^{N}]$, where $\bfx^{n} \in \mathbb{R}^{K \times C}$ and $\bfe^{n} \in \mathbb{R}^{K \times C_e}$ denote the $n$-th motion and expression blocks, respectively.
The waveform is segmented analogously into $N$ blocks of length $K_a$ (\ie, $\bfa = [\bfa^{1}, \ldots, \bfa^{N}]$), where $\bfa^{n} \in \mathbb{R}^{K_a}$ denotes the $n$-th speech segment.

\paragraph{Flow Matching.}
Our method builds on FM~\cite{lipman2023flow,liu2023flow}, a prevailing generative paradigm. 
It constructs a probability path by linearly interpolating between clean motion $\bfx_0$ and noise $\bfz$, governed by time $t$:
\begin{equation}
    \bfx_t = (1- t)\bfx_0 + t \bfz, \quad \bfz \sgaussian, \quad t \in [0,1],
    \label{eq:flow_matching_inter}
\end{equation}
where $\mathcal{N}$ denotes the Gaussian distribution.
Typically, FM learns the path velocity and samples by solving the corresponding probability flow ordinary differential equation~\cite{lipman2023flow,liu2023flow} of this path.

\subsection{Past-as-State, Present-as-Attention Module}
\label{sec:paspa_block}
\begin{wrapfigure}{r}{0.45\linewidth} 
    \centering
    \vspace{-12pt} 
    \includegraphics[width=\linewidth]{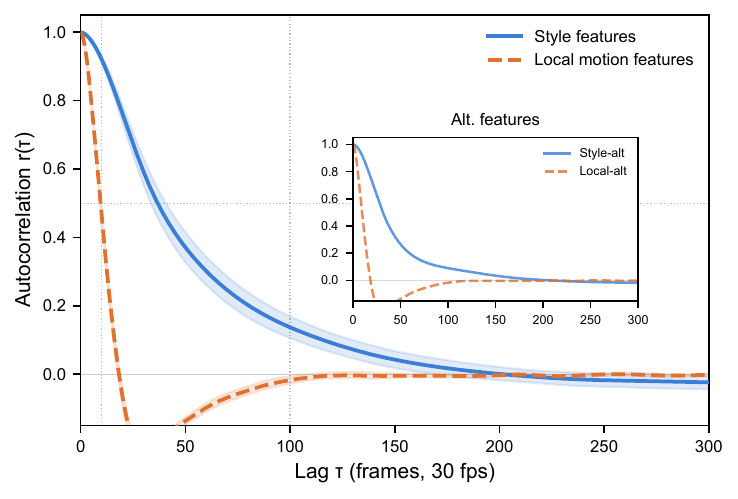}
    \vspace{-10pt} 
    \caption{
        Autocorrelation $r(\tau)$ on BEAT2, averaged across speakers with 95\%  confidence intervals (shaded). 
        Style features remain correlated beyond 100 frames, whereas local features decay rapidly. 
        %
        Inset: raw windowed pose statistics. 
        See \cref{app:evidence_for_asymmetry} for more details.
    }
    \label{fig:acf}
    \vspace{-15pt}
\end{wrapfigure}
In segment-by-segment motion generation~\cite{liu2024emage,zhang2026mitigating}, past history and the current block serve different roles. 
%
Empirically, style cues remain correlated over long timescales, whereas local motion dynamics decay rapidly (see \cref{fig:acf}).
Motivated by this asymmetry, PASPA (see \cref{fig:network}) decouples their processing: an inter-block persistent state compactly summarizes past history, while an intra-block bidirectional attention models the current block via dense interaction.

\subsubsection{Inter-Block Persistent State.}
\label{sec:inter_block}
\paragraph{Why Persistent State for the Past?}
Past history carries stable style and long-range tendencies.
Such information is abstract and sparse: its effective content lies in a few persistent components, and token-level attention would waste capacity on transient fluctuations.
%
Thus, the past should be selectively condensed into a persistent state, not indiscriminately preserved.
%
%
We achieve this via Gated DeltaNet (GDN)~\cite{yang2025gated},
whose gating~\cite{yang2024gated} and delta-rule~\cite{widrow1988adaptive,schlag2021linear} updates suppress fluctuations and accumulate informative trends, yielding a strong inductive bias for history propagation in our data-limited setting.

\begin{figure}[t] 
    \centering
    \includegraphics[width=\linewidth]{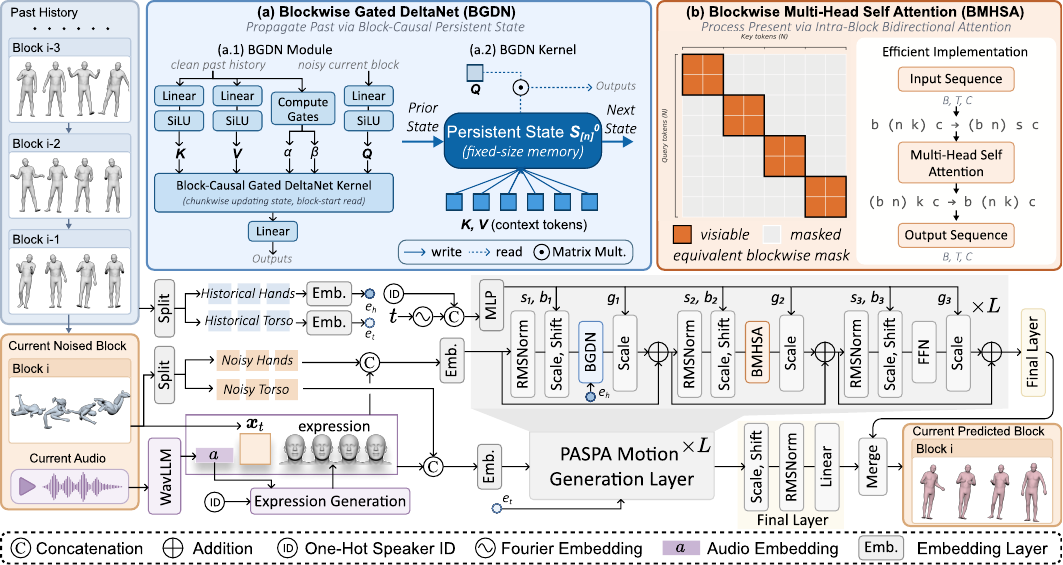}
    \caption{
    Network architecture of PASPA.
    (a) \textbf{BGDN} propagates history through a persistent state $\bfS_{[n]}^0$ via a block-causal GDN kernel.
    (b) \textbf{BMHSA} captures local dynamics via efficient block-diagonal bidirectional attention.
    The overall network (bottom) sequentially applies BGDN and BMHSA. 
    \label{fig:network}
    }
    \vspace{-2mm}
\end{figure}

\paragraph{Blockwise Gated DeltaNet (BGDN).}
GDN is token-recurrent, but our generation proceeds block by block.
%
We thus propose a blockwise GDN (BGDN, \cref{fig:network}a), driven by a tailored Block-Causal GDN kernel.
Inside the BGDN module (\cref{fig:network} (a.1)), we take the noisy current block as input and the encoded clean history as the context.
%
Following GDN, we derive queries $\bfq_i \in \mathbb{R}^{d_k}$ from the input, with keys $\bfk_i \in \mathbb{R}^{d_k}$, values $\bfv_i \in \mathbb{R}^{d_v}$, and gates $\alpha_i, \beta_i \in (0,1)$ from the context, omitting GDN's temporal mixing to prevent cross-block leakage.
%
Within the kernel, these variables update a persistent state $\bfS_i \in \mathbb{R}^{d_v \times d_k}$ and yield outputs $\bfo \in \mathbb{R}^{d_v}$.
$\bfo$ are then projected to produce final outputs.

Inside the kernel (\cref{fig:network} (a.2)), we retain GDN's standard state update to accumulate histories:
\begin{equation}
    \bfS_i = \bfS_{i-1}\lb
        \alpha_i\lb\bfI - \beta_i\bfk_i\bfk_i^{\intercal}\rb
    \rb + \beta_i \bfv_i \bfk_i^{\intercal},
    \label{eq:gdn_write}
\end{equation}
%
where $\alpha_i$ adaptively decays irrelevant fluctuations, and $\beta_i$ governs the delta-rule update for selective information injection. 
However, applying GDN's dynamic read mechanism to blockwise generation would cause intra-block future leakage. 
%
We therefore restrict reads to the frozen block-initial state:
\begin{equation}
    \text{GDN: } \bfo_i={\bfS_i} \bfq_i,
    \qquad
    \text{BGDN: } \bfo_{i}={\bfS_{[n]}^0} \bfq_{i}
    \quad \forall (n-1)K < i \le nK,
    \label{eq:gdn_and_bgdn_read}
\end{equation}
where $\bfS_{[n]}^r$ ($0 \le r \le K$) denotes the state after processing the $r$-th token of block $n$. 
Instead of reading the continuously updated state $\bfS_i$, all queries within block $n$ exclusively read from the block-start state $\bfS_{[n]}^0$ (also the previous block's final state $\bfS_{[n-1]}^K$).
This formulation mathematically guarantees that each block only conditions on strict past history, ensuring the blockwise causality.

\paragraph{Efficient Implementation.}
GDN admits an efficient chunkwise parallel form~\cite{yang2025gated,yang2024fla} (detailed in \cref{app:detailed_parallel_form}). 
Since we leave the state update of GDN unchanged, its parallel computation requires no modification. 
%
The read, however, must be adapted to enforce block-causality. 
Specifically, the computational chunk size is fixed to the generation block size $K$, and all tokens in block $n$ exclusively share the block-start state $\bfS_{[n]}^0$ (\ie, the previous block's final state $\bfS_{[n-1]}^K$). 
This yields:
\begin{equation}
    \bfO_{[n]}^{\mathrm{BGDN}}
    =
    \bfQ_{[n]}
    \left(\bfS_{[n]}^0\right)^{\intercal}
    \in \bbR^{K\times d_v},
    \qquad
    \bfQ_{[n]}\in\bbR^{K\times d_k},
    \quad
    \bfS_{[n]}^0=\bfS_{[n-1]}^K\in\bbR^{d_v\times d_k}.
    \label{eq:bgdn_parallel_read}
\end{equation}
This reduction to a dense matrix multiplication makes the read highly amenable to Tensor Core acceleration, for which we implement a custom Triton~\cite{tillet2019triton} kernel based on FLA~\cite{yang2024fla}.

\subsubsection{Intra-Block Bidirectional Attention.}
\label{sec:intra_block}
\paragraph{Why Bidirectional Attention for the Present?}
The current block carries fast-changing dynamics, such as velocities and fine-grained articulations, which demand dense modeling of local motion details.
%
Correspondingly, BGDN shares a common block-start state, leaving token-to-token interactions within the current block unmodeled.
%
We assign this role to a dedicated bidirectional self-attention module~\cite{vaswani2017attention} rather than a causal one: 
although frames within a block follow a micro-level temporal order, each token still benefits from both preceding and subsequent local context within the segment.

\paragraph{Blockwise Multi-Head Self-Attention (BMHSA).}
Applying bidirectional self-attention~\cite{vaswani2017attention} across the entire sequence naively would break the block-level semi-autoregression.
We therefore propose BMHSA, which permits bidirectional interaction within each block while strictly forbidding any cross-block interaction.
Given an input sequence $\bfX \in \bbR^{NK \times d}$, the output is computed as follows:
\begin{equation}
    \bfO = \mathrm{Softmax}\!\left(\frac{\bfQ \bfK^T}{\sqrt{d}} + \bfB^{\text{blk}}\right) \bfV \bfW_O,
    \quad
    \bfB^{\text{blk}}_{ij} = 
    \begin{cases} 
    0 & \lceil i/K \rceil = \lceil j/K \rceil \\ 
    -\infty & \text{otherwise} 
    \end{cases},
    \label{eq:bmhsa_attn}
\end{equation}
where $i, j \in \{1, \ldots, NK\}$ are token indexes, $\bfB^{\text{blk}} \in \{0, -\infty\}^{NK \times NK}$ is the block-diagonal mask (see \cref{fig:network}b).
We compute the query, key, and value matrices $\bfQ,\bfK,\bfV \in \bbR^{NK\times d}$ as $\bfQ=\bfX\bfW_Q$, $\bfK=\bfX\bfW_K$, and $\bfV=\bfX\bfW_V$, respectively, with rotary position embedding (RoPE)~\cite{su2024roformer} applied to $\bfQ$ and $\bfK$.
Here, $\bfW_Q,\bfW_K,\bfW_V,\bfW_O \in \bbR^{d \times d}$ are projection matrices, with $\bfW_O$ used for output projection.
%
The formulation above is single-head, with the multi-head extension following~\cite{vaswani2017attention}.

\paragraph{Efficient Implementation.}
Rather than materializing $\bfB$, we reshape $\bfX$ from $\bbR^{NK \times d}$ to $\bbR^{N \times K \times d}$ and apply standard bidirectional attention, treating the block index as batch (see \cref{fig:network}b).
This yields the same output as \cref{eq:bmhsa_attn} while reducing the complexity from $\mathcal{O}((NK)^2 d)$ to $\mathcal{O}(N K^2 d)$. 

\subsubsection{Network Architecture.}
\label{sec:architeture}
We instantiate our PASPA layer following the DiT design~\cite{peebles2023scalable} (see \cref{fig:network}) with two modifications.
Structurally, we replace the self-attention sublayer with two sequential sublayers: BGDN and BMHSA. 
For adaptive modulation, we replace AdaLN with AdaRMSNorm~\cite{crowson2024scalable}, where the modulation parameters are predicted from the Fourier-embedded timestep $t$~\cite{tancik2020fourier} and the speaker ID embedding. 
Our architecture follows GlobalDiff~\cite{zhang2026mitigating}, but uses our PASPA layer.
%
%
%
We first \added{use a frozen} audio-to-expression network to obtain expressions from WavLM features~\cite{chen2022wavlm} and speaker ID.
%
Motions are represented in global rotation space and split into torso (with translation) and hands branches.
%
%
%
For each branch, the noisy stream embeds body-part motion with full noisy motion, audio features, and expressions as tokens for the stacked PASPA layers, while the history stream encodes clean past motion into each BGDN module ($\bfe_h, \bfe_t$ in \cref{fig:network}).
The branch outputs are merged to the full motion.

\subsection{Blockwise Flow Matching}
\label{sec:blockwise_flow_matching}

Inspired by recent block diffusion paradigms~\cite{arriola2025block,zhang2025blockvid}, we formalize segment-by-segment motion generation as semi-autoregressive process with block-causal factorization
$
    p(\bfx \mid \bfa, \bfe) = \prod_{n=1}^{N} p(\bfx^n \mid \bfx^{<n}, \bfa^n, \bfe^n)
$.
Here, $\bfx^{<n} = [\bfx^1, \ldots, \bfx^{n-1}]$ provides clean history, $\bfx^n$ is the current denoising target, and $\bfa^n, \bfe^n$ are the aligned speech and expression condition of block $n$.
We precompute $\bfe^n$ from $\bfa^n$ via an audio-to-expression network (\cref{sec:architeture}).
%
This process applies inter-block autoregression for long-horizon causality and intra-block bidirectional modeling for dense local dynamics.

\paragraph{Multi-Block Supervision.}
To realize the semi-autoregressive factorization, we instantiate the FM interpolation in \cref{eq:flow_matching_inter} for each block, allowing every block to carry its own timestep $t^n \in [0,1]$:
\begin{equation}
    \bfx_{t^n}^n = (1-t^n)\bfx_0^n + t^n\bfz^n, \qquad \bfz^n \sgaussian, \qquad \forall n=1,\ldots,N,
    \label{eq:bfm_path}
\end{equation}
where $\bfx_0^n$ is the $n$-th clean block.
Following common practice in motion generation~\cite{zhang2026mitigating,tevet2022human}, we train the network $\mathcal{F}_\theta$ to directly predict each clean block $\bfx_0^n$, rather than the velocity in FM~\cite{lipman2023flow}. 
%
%
The block-causal $\mathcal{F}_\theta$ performs dense parallel denoising over the entire sequence in a single forward pass:
\begin{equation}
    \hat{\bfx}_0
    =
    [\hat{\bfx}_0^1,\hat{\bfx}_0^2,\ldots,\hat{\bfx}_0^N]^{\intercal}
    =
    \mathcal{F}_\theta([\bfx_{t^1}^1,\bfx_{t^2}^2, \ldots, \bfx_{t^N}^N]^{\intercal}, \bfx_0, \bft, \bfa),
    \quad 
    \bft = [t^1, t^2, \ldots, t^N]^{\intercal},
    \label{eq:bfm_network}
\end{equation}
where $\bfa$ is the aligned speech and $\bfx_0 = [\bfx_0^1, \ldots, \bfx_0^N]^{\intercal}$ is encoded as ground-truth history following teacher forcing~\cite{ning2024elucidating,schmidt2019generalization}. 
BGDN routes only $\bfx_0^{<n}$ into block $n$'s prediction to preserve block-causality (\cref{sec:inter_block}).
%
Unlike per-block supervision~\cite{zhang2025semtalk,zhang2026mitigating,liu2024emage}, each iteration accumulates loss over all blocks, yielding $N\times$ denser supervision.
%
More critically, heterogeneous block timesteps expose the model to diverse current-block noise levels at each history length, improving rollout robustness.
%
Notably, we enforce $t^1 \le t^2 \le \cdots \le t^N$ to mirror the inference-time noise profile, and replace the typical uniform timestep distribution with a logit-normal one~\cite{esser2024scaling}.
Both are supported by ablations in \cref{tab:ablation_train}.

\paragraph{Loss Function.}
We supervise the predicted $\hat{\bfx}_0$ in \cref{eq:bfm_network} with the following overall objective\footnote{With a slight abuse of notation, we use $\mathcal{L}$ to denote both the scalar loss term and the corresponding loss function.}:
\begin{align}
    &
    \mathcal{L}
    =
    \mathbb{E}_{\bfx_0,\bfa,\bfz,\bft}
    \left[
    \frac{1}{N}
    \sum_{n=1}^{N}
    \lsb
        e^{-u(t^n)} \mathcal{L}_{\mathrm{rot}}^n
            +
        \mathsf{sg}\lb e^{-u(t^n)} \rb   \lb\mathcal{L}_{\mathrm{joint}}^n
            +
        \mathcal{L}_{\mathrm{struct}}^n \rb
        +
        u(t^n)
    \rsb
    +
    \mathcal{L}_{\mathrm{bd}}
    \right],
    \label{eq:bfm_total_loss}
    \\
    &
    \mathcal{L}_{\mathrm{bd}}
    =
    \frac{1}{N-1}
    \sum_{n=1}^{N-1}
    \left[
        \mathcal{L}_{\mathrm{rot}}\!\left(\hat{\bfb}_n,\bfb_n\right)
        +
        \mathcal{L}_{\mathrm{vel}}\!\left(\hat{\bfb}_n,\bfb_n\right)
        +
        \mathcal{L}_{\mathrm{acc}}\!\left(\hat{\bfb}_n,\bfb_n\right)
    \right].
    \label{eq:bfm_bd_loss}
\end{align}
Here, $\mathcal{L}_{\mathrm{rot}}^n$ is the MSE loss on rot6d~\cite{zhou2019continuity} and translations, while $\mathcal{L}_{\mathrm{joint}}^n$ and $\mathcal{L}_{\mathrm{struct}}^n$ are forward kinematics (FK)-based losses on joint positions and skeletal structures adopted from GlobalDiff~\cite{zhang2026mitigating}.
To adaptively mitigate optimization conflicts across blocks at varying noise levels, we train a small MLP $u(\cdot)$ to predict per-timestep weights~\cite{karras2024analyzing} rather than manual tuning.
We apply a stop-gradient ($\mathsf{sg}$) to prevent the fluctuations of FK-based losses from destabilizing $u(\cdot)$.
The boundary loss $\mathcal{L}_{\mathrm{bd}}$ regularizes cross-block continuity via velocity and acceleration matching losses (\ie, $\mathcal{L}_{\mathrm{vel}}, \mathcal{L}_{\mathrm{acc}}$) on the transition windows $\hat{\bfb}_n$ and $\bfb_n$ of $\hat{\bfx}_0^n$ and $\bfx_0^n$, and is excluded from the adaptive weighting as each window spans two blocks at different timesteps.
See \cref{app:loss_details} for definitions of $\mathcal{L}_{\mathrm{rot}}, \mathcal{L}_{\mathrm{joint}}, \mathcal{L}_{\mathrm{struct}}, \mathcal{L}_{\mathrm{vel}}, \mathcal{L}_{\mathrm{acc}}, \hat{\bfb}_n, \bfb_n$.

\begin{algorithm}[t]
\caption{Inference with hybrid CFG, where $\bfc$ denotes the joint condition of speech and expression.}
\label{alg:rollout_cfg}
\small
\begin{algorithmic}[1]
\Require 
Condition $\bfc^{1:N}$; 
Timesteps $\{t_k\}_{k=1}^M$; 
CFG scales $w_\mathrm{cond}(t), w_\mathrm{hist}(t)$; 
seed $(\bfx_{\mathrm{seed}}, \bfc_{\mathrm{seed}})$ 
\State $\bfS \gets \mathrm{Prefill}(\bfS,\, \bfx_{\mathrm{seed}},\, \bfc_{\mathrm{seed}})$ \textbf{if} $(\bfx_{\mathrm{seed}}, \bfc_{\mathrm{seed}})$ is not $\emptyset$ \textbf{else} $\boldsymbol{0}$ \Comment{initialize GDN state}
\For{$n = 1, 2, \ldots, N$} \Comment{iterate for $N$ blocks}
    \State $\bfx_{t_1}^n \sgaussian$ 
    \Comment{initial noise from Gaussian distribution}
    \For{$k = 1, 2, \ldots, M{-}1$} \Comment{iterate for $M-1$ denoising steps}
        \State 
        $\hat{\bfx}_0^{n,\emptyset} \gets \mathcal{F}_\theta(\bfx_{t_k}^n, t_k, \bfS_\emptyset, \emptyset)$,
        $\hat{\bfx}_0^{n,\mathrm{c}} \gets \mathcal{F}_\theta(\bfx_{t_k}^n, t_k, \bfS_\emptyset, \bfc^n)$
        \Comment{unconditional and condition-only}
        \State
        $\hat{\bfx}_0^{n,\mathrm{h}} \gets \mathcal{F}_\theta(\bfx_{t_k}^n, t_k, \bfS, \emptyset)$,
        $\hat{\bfx}_0^{n} \gets \mathcal{F}_\theta(\bfx_{t_k}^n, t_k, \bfS, \bfc^n)$
        \Comment{history-only and full}
        \State
        $\begin{aligned}
        \hat{\bfx}_0^n \gets &\hat{\bfx}_0^{n,\emptyset} 
            + w_\mathrm{cond}(t)(\hat{\bfx}_0^{n,c} - \hat{\bfx}_0^{n,\emptyset}) 
            + w_\mathrm{hist}(t)(\hat{\bfx}_0^{n,h} - \hat{\bfx}_0^{n,\emptyset}) \\
        & + w_\mathrm{cond}(t) w_\mathrm{hist}(t)(\hat{\bfx}_0^{n} - \hat{\bfx}_0^{n,c} - \hat{\bfx}_0^{n,h} + \hat{\bfx}_0^{n,\emptyset})
        \end{aligned}$
        \Comment{hybrid CFG}
        \State $\bfx_{t_{k+1}}^n \gets \mathrm{SamplingStep}(\bfx_{t_k}^n,\, \hat{\bfx}_0^n,\, t_k,\, t_{k+1})$
        \Comment{take a sampling step for denoising}
    \EndFor
    \State $\bfS \gets \mathrm{Prefill}(\bfS,\, \hat{\bfx}_0^n,\, \bfc^n)$ \Comment{write block into $\bfS$}
\EndFor
\State \Return $\hat{\bfx}_0^{1:N}$ \Comment{concatenate all generated motion segments into the final result}
\end{algorithmic}
\end{algorithm}

\subsection{Inference for Long-Horizon Rollout}
\label{sec:inference}

\paragraph{Prefill-Decode Rollout with State Caching.}
For long-horizon inference, PASPA employs a prefill-decode rollout (see \cref{alg:rollout_cfg}).
\emph{Prefill} only encodes the clean segment to update the BGDN state, bypassing the denoising sublayers.
%
Invoked after each block iteration (line~8), it integrates the freshly denoised block into the state.
%
\emph{Decode} (lines~4-7) treats $\bfS$ strictly as a read-only condition. 
%
It iteratively denoises the current block (line~7), evaluating $\mathcal{F}_\theta$ under varying conditions for hybrid CFG (detailed below).
%
%
Throughout the rollout, the fixed-size $\bfS$ is the sole persistent cache, initialized by prefilling seed $\bfx_{\mathrm{seed}}$ and $\bfa_{\mathrm{seed}}$ (or zeros) and reused across all $N$ blocks. 
Unlike linearly growing token caches~\cite{vaswani2017attention,kwon2023efficient}, $\bfS$ guarantees an $\mathcal{O}(1)$ query complexity per step, regardless of history length.

\paragraph{Hybrid Classifier-Free Guidance.}
Standard CFG~\cite{ho2022classifier} modulates only the condition $\bfc^n$, leaving historical dependence uncontrolled. 
%
%
Motivated by \cite{song2025history}, we introduce a learnable null state $\bfS_\emptyset$ for hybrid guidance via four predictions (\cref{alg:rollout_cfg}, lines~5-6): unconditional $(\bfS_\emptyset, \emptyset)$, condition-only $(\bfS_\emptyset, \bfc^n)$, history-only $(\bfS, \emptyset)$, and full $(\bfS, \bfc^n)$.
They are combined via a multilinear composition (\cref{alg:rollout_cfg}, line~7) that incorporates marginal and second-order joint terms, empirically outperforming simpler alternatives (see \cref{tab:hybrid_cfg_mode}).
%
To prevent distortions at noise extremes~\cite{kynkaanniemi2024applying}, scales $w_\mathrm{cond}(t)$ and $w_\mathrm{hist}(t)$ are time-dependent, activating only during mid-noise intervals $[0.1,0.6]$ (otherwise $1$).
%
%
%
To train Hybrid CFG, we independently drop $\bfc^n$ with probability $10\%$ and sample $\bfS$ as unchanged, null $\bfS_\emptyset$, or interpolated $\alpha\bfS + (1{-}\alpha)\bfS_\emptyset$ with probabilities $70\%/10\%/20\%$, where $\alpha$ is uniformly sampled from $[0,1]$.
This interpolation enables continuous inference-time control of history strength.
\section{Experiments}
\label{sec:experiments}

\paragraph{Setup.}
\added{We evaluate on BEAT2~\cite{liu2024emage} using its official splits, and on SHOW~\cite{yi2023generating} following the evaluation protocol of DiffSHEG~\cite{chen2024diffsheg}.}
\added{Following~\cite{mughal2025retrieving}, BEAT2 comparisons use both the ``Scott'' subset (\textbf{1 Speaker}) and the full test set (\textbf{All Speakers}), while ablations use the \textbf{1 Speaker} setting.}
PASPA is optimized via Muon~\cite{liu2025muon,jordan2024muon} with auxiliary AdamW~\cite{kingma2014adam} (training details in \cref{app:imple_main_results}).
\added{We report FGD~\cite{yoon2020speech} (the primary quality indicator) and Diversity (Div)~\cite{li2021audio2gestures} on both datasets, with BC~\cite{li2021ai} and
face MSE~\cite{yang2023diffusestylegesture} on BEAT2, and Beat Alignment (BA) on SHOW.}
Here, $\uparrow$/$\downarrow$ indicate that higher/lower values are better.
\added{For readability, BEAT2 FGD and MSE are multiplied by $10$ and $10^{8}$, respectively; SHOW metrics are reported without rescaling.}
In all tables, gray rows mark our final setting; \best{bold} and \sbest{underlined} denote the best and second-best results.
See \cref{app:dataset_information,app:implementation_details,app:additional_experimental_results} for details.

\begin{table*}[t]
\centering
\renewcommand{\arraystretch}{1.0}
\setlength{\tabcolsep}{4pt}
\newcolumntype{C}{>{\centering\arraybackslash}X}
\caption{
    Comparison with state-of-the-art (SOTA) methods on the BEAT2 dataset. 
    \label{tab:comparison}
}
\vspace{-2mm}
\begin{tabularx}{\textwidth}{l *{8}{C}}
\toprule
\multirow{2}{*}{\textbf{Method}} 
& \multicolumn{4}{c}{\textbf{1 Speaker}} 
& \multicolumn{4}{c}{\textbf{All Speakers}} \\
\cmidrule(lr){2-5} \cmidrule(lr){6-9}
 & FGD\,$\downarrow$ & BC\,$\uparrow$ & Div\,$\uparrow$ & MSE\,$\downarrow$ 
 & FGD\,$\downarrow$ & BC\,$\uparrow$ & Div\,$\uparrow$ & MSE\,$\downarrow$ \\
\midrule
%
CaMN~\cite{liu2022beat}   
& 6.64 & 0.677 & 10.86 & --     
& 5.12 & 0.200 & 5.58 & -- \\
DSG~\cite{yang2023diffusestylegesture}
& 8.81     & 0.724     & 11.49     & --        
& 11.74     & \sbest{0.734}      & \sbest{11.12}     & --      \\
%
%
%
EMAGE~\cite{liu2024emage}
& 5.51 & 0.772 & 13.06 & 7.680 
& 6.92  & 0.284 & 6.06 & 6.908 \\
DiffSHEG~\cite{chen2024diffsheg}
& 8.99 & 0.714 & 11.91 & 7.665 
& --    & --    & --    & --     \\
MambaTalk~\cite{xu2024mambatalk}
& 5.37 & \sbest{0.781} & 13.05 & 6.289  
& -- & -- & -- & --  \\
HoloGest~\cite{cheng2025hologest}
& 5.34 & \best{0.795} & \best{14.15} & -- 
& 6.46 & \best{0.803} & \best{13.53} & -- \\
RAG-GESTURE~\cite{mughal2025retrieving}
& 8.08 & 0.734 & 11.97 & -- 
& 4.87 & 0.514 & 9.94 & -- \\
SemTalk~\cite{zhang2025semtalk}
& 4.28 & 0.777 & 12.91 & \sbest{6.153} 
& --    & --    & --    & --    \\
GestureLSM~\cite{liu2025gesturelsm}
& \sbest{4.09} & 0.714 & 13.24 & -- 
& --    & --    & --    & --      \\
GlobalDiff~\cite{zhang2026mitigating}
& 4.78 & 0.705 & 13.13  & 6.330   
& \sbest{2.63} & 0.404 & 8.24   & \sbest{4.144} \\
\midrule
\rowcolor{graybg}
Ours 
& \best{3.71} & 0.746 & \sbest{13.38} & \best{5.768} 
& \best{2.33} & 0.645 & 9.42 & \best{3.793} \\
\bottomrule
\end{tabularx}
\end{table*}

\begin{figure}[t] 
    \centering
    \includegraphics[width=\linewidth]{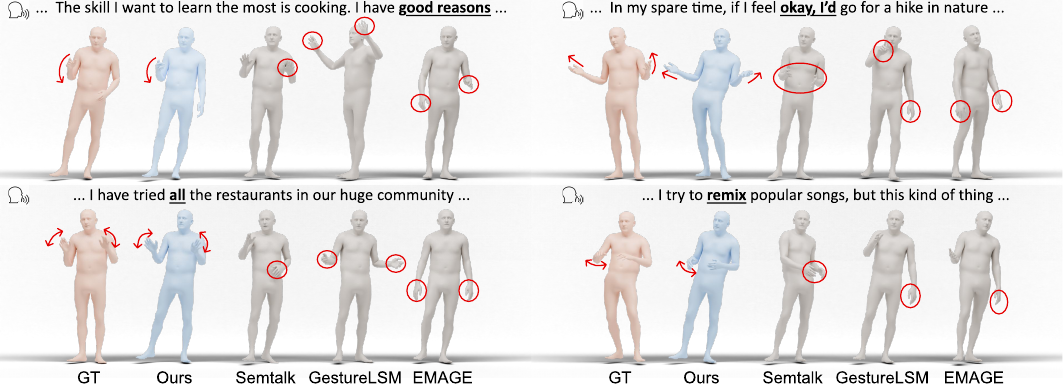}
    \caption{
         Qualitative comparison.
         PASPA consistently matches the GT in gesture direction (red arrows), amplitude, and speech alignment.
         Baselines exhibit various failure modes (red circles): unnatural hand positions, hesitant or mismatched gestures, and missing semantic response.
        \label{fig:visual_results}
    }
\end{figure}

\subsection{Main Results}
\label{sec:comparison}

\paragraph{\added{Results on BEAT2.}}
\Cref{tab:comparison} compares PASPA with prior methods \added{on BEAT2}.
%
In the 1 Speaker setting, PASPA achieves the best FGD (the primary quality metric) and MSE, indicating superior generation quality.
%
It \added{also} obtains the second-best Div and competitive BC, reflecting favorable diversity and speech-motion alignment.
In the All Speakers setting, PASPA achieves the best FGD and MSE, showing \added{its} stronger multi-speaker generalization.
%
Compared with the FM-based GlobalDiff~\cite{zhang2026mitigating}, PASPA improves all metrics, suggesting the advantage of persistent-state history modeling for FM-based co-speech motion generation.
%
\Cref{fig:visual_results} also qualitatively shows \added{that} PASPA produces more natural gestures with better semantic alignment \added{to speech} than \added{the} baselines.
%
%
%

\begin{figure}[t]
    \centering
    \captionsetup{font=small,skip=4pt,singlelinecheck=false}

    \begin{minipage}[t]{0.45\linewidth}
        \vspace{0pt}
        \centering
        \captionsetup{type=table,position=top,skip=2pt}
        \caption{\added{
        Comparison on the SHOW dataset. 
        }}
        \label{tab:show_comparison}

        \small
        \ifmarkadditions\color{blue}\fi
        \ifdefined\hypersetup\hypersetup{hidelinks}\fi
        \arrayrulecolor{black}
        \renewcommand{\arraystretch}{1.08}
        \setlength{\tabcolsep}{2pt}
        \begin{tabularx}{\linewidth}{l*{3}{>{\centering\arraybackslash}X}}
            \toprule
            Method & FGD\,$\downarrow$ & BA\,$\uparrow$ & Div\,$\uparrow$ \\
            \midrule
            GT & -- & 0.867 & 0.834 \\
            \midrule
            LS3DCG~\cite{habibie2021learning}
              & 0.00478 & \best{0.947} & 0.645 \\
            TalkSHOW~\cite{yi2023generating}
              & 0.00328 & 0.872 & 0.711 \\
            DiffSHEG~\cite{chen2024diffsheg}
              & 0.00271 & \sbest{0.902} & 0.810 \\
            GlobalDiff~\cite{zhang2026mitigating}
              & \sbest{0.00198} & 0.870 & \best{1.180} \\
            \rowcolor{graybg}
            PASPA & \best{0.00111} & 0.876 & \sbest{1.043} \\
            \bottomrule
        \end{tabularx}
    \end{minipage}\hfill%
    \begin{minipage}[t]{0.53\linewidth}
        \vspace{0pt}
        \centering
        \color{black}
        \setlength{\parskip}{0pt}
        \captionsetup{type=figure,position=bottom,skip=2pt}
    
        \graphicspath{{assets/}{../assets/}}
        \includegraphics[width=\linewidth]
            {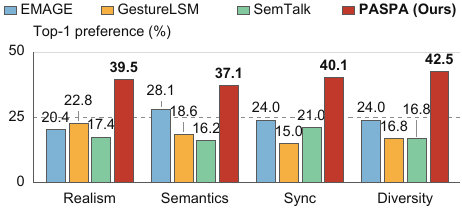}\par
        \nointerlineskip
        \caption{\added{User-study top-$1$ win rates (\%; dashed: $25\%$).}}
        \label{fig:user_study_winrate}
    \end{minipage}
\end{figure}

\added{We further conduct a blind ranking study on BEAT2 with $18$ participants, yielding $167$ valid ranking groups.}
\added{Participants rank four anonymized methods on $20$-second clips across four criteria, with randomized method order and a ground-truth reference.}
\added{The four criteria are motion realism, semantic consistency with speech, motion-speech synchronization, and motion diversity.}
\added{PASPA achieves the highest top-$1$ win rate across all four criteria (see \cref{fig:user_study_winrate}), with $39.8\%$ overall, further supporting these findings.}
\added{See \cref{app:imple_user_study,app:user_study} for study details and the supplementary material for video demos.}

\paragraph{\added{Results on SHOW.}}
\added{To complement our primary evaluation, we evaluate PASPA on SHOW using the same training recipe as on BEAT2 without dataset-specific hyperparameter tuning.}
\added{As shown in \cref{tab:show_comparison}, PASPA achieves the lowest FGD, our primary generation-quality metric, at $0.00111$, a $43.9\%$ reduction from our FM-based baseline GlobalDiff, which achieves an FGD of $0.00198$.}
\added{PASPA yields slightly higher BA than GlobalDiff, while its Div remains second best.}
\added{These results provide complementary evidence of PASPA's effectiveness beyond BEAT2 in co-speech motion generation.}

\subsection{Ablations and Analysis}
\label{sec:ablation}

\added{As outlined in \cref{sec:methodology}, PASPA's architecture realizes the past--present asymmetry through inter-block persistent-state modeling and intra-block bidirectional attention.}
\added{Weakly monotonic timesteps and adaptive weighting stabilize multi-block learning, while Hybrid CFG refines generation quality at inference time and prefill-decode separation accelerates generation.}
\added{The following ablations and analyses examine these complementary roles and their effects on generation quality and efficiency.}

\begin{table}[t]
    \centering 
    \renewcommand{\arraystretch}{1.0} 
    \setlength{\tabcolsep}{3pt} 
    \small
    \newcolumntype{Y}{>{\centering\arraybackslash}X}
    
    
    \begin{minipage}[t]{0.52\linewidth}
    \centering
    \caption{
        Ablation on past-present modeling.
        \textbf{Past}/\textbf{Curr.}: past/current block modeling. 
        $\text{Attn}_{K}$ ($\text{Attn}_{\text{all}}$): attention over $K$-frame (full) history.
        \label{tab:ablation_asym}
    }
    \vspace{4pt}
    \begin{tabularx}{\linewidth}{@{} l l *{4}{Y} @{}}
        \toprule
        \textbf{Past} & \textbf{Curr.}
        & \textbf{FGD}$\downarrow$ & \textbf{BC}$\uparrow$
        & \textbf{Div}$\uparrow$ & \textbf{MSE}$\downarrow$ \\
        \midrule
        --                           & Attn  
        & 7.065 & 0.743 & 13.16 & 5.727 \\
        
        $\text{Attn}_{8}$            & Attn  
        & 6.051 & 0.721 & 12.14 & 5.765 \\
        
        
        $\text{Attn}_{\text{all}}$   & Attn  
        & 5.710 & 0.744 & 11.14 & 5.696 \\
        
        State                        & State 
        & 7.652 & 0.785 & 11.19 & 6.264 \\
        
        \rowcolor{graybg}
        State               & Attn
        & 5.216 & 0.734 & 14.26 & 5.761 \\
        \bottomrule
    \end{tabularx}
\end{minipage}

    \hfill
    \begin{minipage}[t]{0.45\linewidth} 
    \centering
    \caption{
        Ablation on training build-up. 
        Each ``$+$'' row inherits all components above.
        \label{tab:ablation_train}
    }
    \begin{tabularx}{\linewidth}{@{} l *{4}{Y} @{}}
        \toprule
        \textbf{Variant}
        & \textbf{FGD}$\downarrow$ & \textbf{BC}$\uparrow$
        & \textbf{Div}$\uparrow$ & \textbf{MSE}$\downarrow$ \\
        \midrule
        \emph{Single}             
        & 7.410 & 0.744 & 11.95 & 6.341 \\
        \emph{Multi}              
        & 8.456 & \best{0.747} & 12.58 & 6.090 \\
        \midrule
        \hspace{6pt}$+$ Mono   
        & 6.346 & 0.738 & 13.78 & 6.385 \\
        
        \hspace{6pt}$+$ AdaW 
        & 5.614 & 0.736 & \best{14.93} & \best{5.736} \\
        
        \hspace{6pt}$+$ SG 
        & 5.216 & 0.734 & 14.26 & 5.761 \\
        
        \rowcolor{graybg}
        \hspace{6pt}$+$ LogitN
        & \best{4.450} & 0.739
        & 12.36 & 5.763 \\
        
        \bottomrule
    \end{tabularx}
\end{minipage}
    \vspace{-2mm}
\end{table}

\paragraph{Asymmetric Past-Present Modeling.}
We validate the PASPA principle by varying the past and current modeling under a fixed training recipe in \cref{tab:ablation_asym}. 
Removing history (history branch output set to zero) substantially degrades FGD, confirming the necessity of long-range context.
Increasing token-attention history from $\text{Attn}_{8}$ to $\text{Attn}_{\text{all}}$ improves FGD, but still underperforms PASPA, showing that retaining more raw history does not guarantee effective history utilization.
%
%
\added{Moreover, replacing current-block attention with state-based recurrent modeling (\ie, State+State) yields worse FGD than even the no-history variant.}
\added{This shows that the current block benefits from dense bidirectional interaction and that state-based current-block modeling hurts generation quality.}
%
%
These comparisons validate PASPA's  \emph{past-as-state, present-as-attention} division, which yields the best FGD.

\paragraph{Blockwise Training Formulation.}
We show the build-up ablation of our training recipe in \cref{tab:ablation_train}.
%
Naively switching from single-block (\emph{Single}) to multi-block (\emph{Multi}) supervision degrades FGD despite denser supervision, due to unordered block timesteps and imbalanced losses across blocks and noise levels.
Weakly monotonic timesteps (Mono) and adaptive loss weighting (AdaW; see \cref{eq:bfm_total_loss}) address these two issues, significantly reducing FGD to 5.614. 
Beyond the two critical stabilizers, SG (\ie, $\mathsf{sg}(\cdot)$ in \cref{eq:bfm_total_loss}) reduces FGD to 5.216, and LogitN (\ie, logit-normal time sampling~\cite{esser2024scaling}) brings it to 4.450.
We use the optimal LogitN parameters in \cref{tab:ablation_train}; see \cref{tab:lognorm_sweep} for the full sweep.

\begin{figure}[t] 
    \centering
    \newcolumntype{Y}{>{\centering\arraybackslash}X}
    \renewcommand{\arraystretch}{1.0}
    \begin{minipage}[t]{0.62\linewidth} 
        \vspace{0pt} 
        \centering
        
        \captionsetup{type=table} 
        \caption{
            CFG ablation. Cond (Hist): condition (history) dropout and guidance. $(w_{\mathrm{cond}}, w_{\mathrm{hist}})$ simplified to $(w_{\mathrm{c}}, w_{\mathrm{h}})$.
        }
        \label{tab:ablation_cfg_main}
        
        \small
        \renewcommand{\arraystretch}{1.18}
        \setlength{\tabcolsep}{3pt} 
        
        \begin{tabularx}{\linewidth}{@{} l l @{} c @{} *{4}{Y} @{}}
            \toprule
            \textbf{Dropout}
            & \textbf{Guidance}
            & {\small $\boldsymbol{\,(w_{\mathrm{c}}, w_{\mathrm{h}})\,}$}
            & \textbf{FGD}\,$\downarrow$
            & \textbf{BC}\,$\uparrow$
            & \textbf{Div}\,$\uparrow$
            & \textbf{MSE}\,$\downarrow$ \\
            \midrule

                       --
            & --
            & $(\text{--},\phantom{0}\text{--})$ 
            & 4.450 & 0.739 & 12.36 & \best{5.763} \\
    
            Cond
            & --
            & $(1.0, \phantom{0}\text{--})$
            & 3.937 & 0.741 & {13.46} & 5.814 \\
    
            Cond
            & Cond
            & $(1.5, \phantom{0}\text{--})$
            & {3.883} & 0.740 & 13.39 & 5.823 \\
    
            Cond+Hist
            & Cond
            & $(1.5, 1.0)$
            & 3.989 & {0.743} & 13.17 & 5.782 \\
    
            Cond+Hist
            & Hybrid$^\dagger$
            & $(1.5, 1.5)$
            & 3.967 & 0.742 & \best{13.76} & 5.800  \\

            \rowcolor{graybg}
            Cond+Hist
            & Hybrid
            & $(1.5, 1.5)$
            & \best{3.532} & \best{0.745} & {13.30} & 5.779  \\
            \bottomrule
            
            \multicolumn{7}{@{}p{\linewidth}@{}}{
                \footnotesize $^\dagger$ We remove the second-order joint term in line~7 of \cref{alg:rollout_cfg}.
            }
        \end{tabularx}
        
        
        \captionsetup{type=table} 
            \caption{
                Long-horizon generation. 
                FGD and Div evaluated on the terminal $10$\,s window across $T \in \{10, 30, 60\}$\,s rollouts.
                \label{tab:long_rollout}
            }
            \begin{tabularx}{\linewidth}{@{} l *{6}{Y} @{}}
                \toprule
                \multirow{2.5}{*}{\textbf{Method}} & 
                \multicolumn{2}{c}{\textbf{[0, 10]s}} & 
                \multicolumn{2}{c}{\textbf{[20, 30]s}} & 
                \multicolumn{2}{c}{\textbf{[50, 60]s}} \\
                
                \cmidrule(lr){2-3} \cmidrule(lr){4-5} \cmidrule(lr){6-7}
                
                & \textbf{FGD}\,$\downarrow$ & \textbf{Div}\,$\uparrow$
                & \textbf{FGD}\,$\downarrow$ & \textbf{Div}\,$\uparrow$
                & \textbf{FGD}\,$\downarrow$ & \textbf{Div}\,$\uparrow$ \\
                \midrule
                
                Short Seed     
                & 4.731         & 6.581          
                & 5.886         & 6.033          
                & 6.721         & 6.081 \\
                Token Attention 
                & \best{4.448}  & 6.289          
                & 5.388         & 6.510          
                & 5.771         & 6.427 \\
                \rowcolor{graybg}
                PASPA          
                & \best{4.448}      & \best{7.217}   
                & \best{5.119}      & \best{7.133}   
                & \best{5.745}      & \best{7.138} \\
                
                \bottomrule
            \end{tabularx}
            
            
            
        
    \end{minipage}
    \hfill
    \begin{minipage}[t]{0.36\linewidth}
        \vspace{0pt} 
        \centering
        \includegraphics[width=\linewidth]{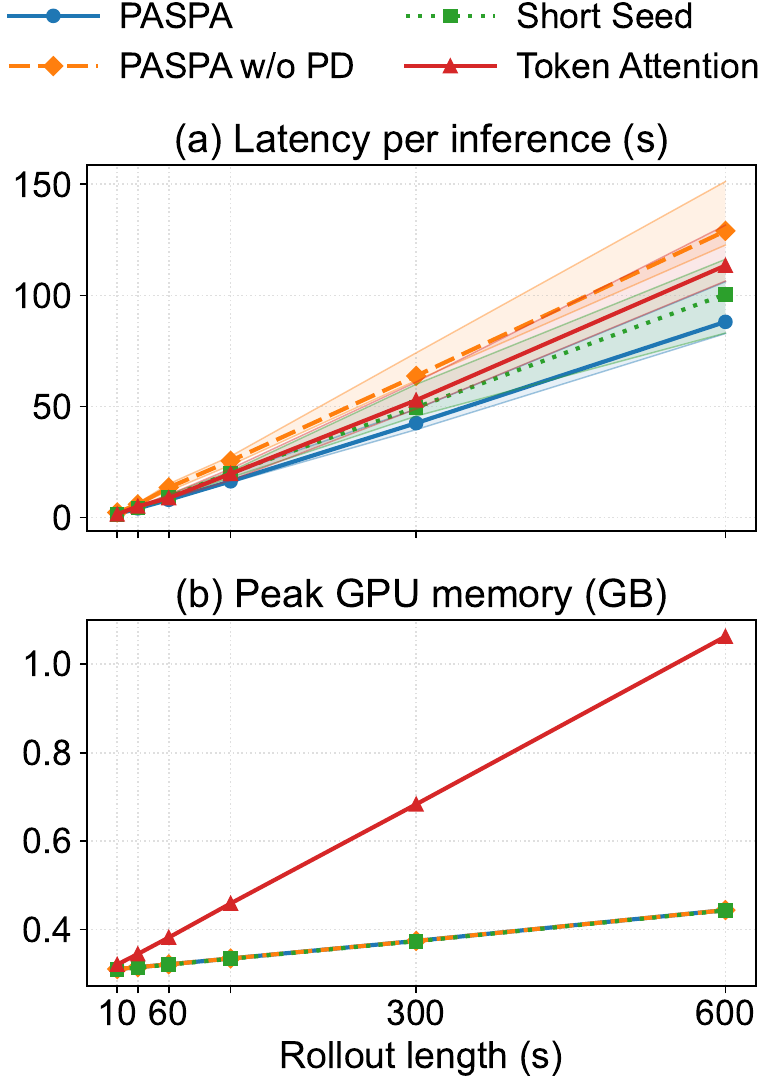}
        \vspace{-4mm}
        \captionsetup{type=figure} 
        \caption{
            Efficiency analysis. 
            Lines: median latency and peak GPU memory. 
            Shaded bands: min-max latency.
           \label{fig:infer_efficiency}
        }
    \end{minipage}
\end{figure}

\paragraph{Classifier-Free Guidance.} 
\Cref{tab:ablation_cfg_main} evaluates the CFG build-up, including training dropout and inference guidance.
%
%
Condition dropout, acting as data augmentation, already reduces FGD from $4.450$ to $3.937$, and condition guidance further improves it to $3.883$.
%
%
%
However, history dropout without guidance slightly degrades FGD, emphasizing that it offers no inherent quality gain alone.
Further adding history guidance without the second-order joint term also remains worse than condition CFG. 
%
Only the full Hybrid CFG substantially reduces FGD to $3.532$, validating the explicit interaction between condition and history guidance.
%
The activation interval is fixed to $t \in [0.1, 0.6]$.
Detailed CFG scale, composition, and activation-interval ablations are in \cref{tab:cfg_interval,tab:cond_cfg_scale,tab:hybrid_cfg_scale,tab:hybrid_cfg_mode}.
\added{An end-to-end comparison with and without CFG under the main-experiment setting is in \cref{tab:cfg_main_results}.}

\paragraph{Long-Rollout Analysis.} 
We quantify long-rollout quality preservation in \cref{tab:long_rollout} (implementation in \cref{app:imple_other_exp}).
\added{We construct controlled baselines: \emph{Short Seed} uses the last $8$ frames, while \emph{Token Attention} attends to all past tokens.}
\added{They represent finite-window and full-history conditioning, respectively, with other settings matched to PASPA.}
\added{We use these controlled variants rather than complete methods to isolate history propagation from method-specific differences (\eg, architecture, losses).}
Although all methods start with reasonable FGD, Short Seed degrades most severely, Token Attention degrades less by retaining explicit history, and PASPA shows the weakest degradation, proving persistent states better preserve long-range tendencies.
%
%
PASPA also achieves the highest diversity (Div) throughout rollout. 
%
\Cref{fig:teaser} shows consistent qualitative trends, supporting PASPA's long-rollout advantage.
\added{See \cref{fig:app_long_rollout_curves} for complementary long-rollout trends, additional metrics, and uncertainty estimates.}

%
%
We further test efficiency \added{without CFG} on synthetic ultra-long rollouts in \cref{fig:infer_efficiency}, where PASPA w/o PD denotes PASPA without prefill-decode separation.
%
PASPA matches Short Seed's peak memory and stays far below Token Attention's, as history is accumulated into a fixed-size state instead of growing explicit tokens.
%
While latency increases with length for all methods, prefill-decode separation is decisive: it accelerates PASPA from the slowest to the fastest.
See \cref{app:imple_other_exp} for details.
%
%

\begin{table}[t]
    \centering
    \caption{
    \added{
        Inference overhead of Hybrid CFG under the final PASPA configuration. 
        %
        %
    }}
    \label{tab:cfg_efficiency_h100}
    \renewcommand{\arraystretch}{1.20}
    \setlength{\tabcolsep}{4.5pt}
    \ifmarkadditions\color{blue}\fi
    \arrayrulecolor{black}
    \begin{tabular}{l*{6}{c}}
        \toprule
        \multirow{2}{*}{\textbf{Guidance}}
        & \multicolumn{2}{c}{\textbf{120\,s (3{,}600 frames)}}
        & \multicolumn{2}{c}{\textbf{300\,s (9{,}000 frames)}}
        & \multicolumn{2}{c}{\textbf{600\,s (18{,}000 frames)}} \\
        \cmidrule(lr){2-3}
        \cmidrule(lr){4-5}
        \cmidrule(lr){6-7}
        & \textbf{Lat. (ms)} & \textbf{Mem. (GB)}
        & \textbf{Lat. (ms)} & \textbf{Mem. (GB)}
        & \textbf{Lat. (ms)} & \textbf{Mem. (GB)} \\
        \midrule

        No CFG
        & 11{,}411 & 0.36
        & 28{,}562 & 0.40
        & 55{,}978 & 0.47 \\

        \rowcolor{graybg}
        Hybrid CFG
        & 12{,}439 & 0.37
        & 29{,}966 & 0.41
        & 61{,}093 & 0.48 \\

        \bottomrule
    \end{tabular}
\end{table}

\paragraph{\added{Hybrid CFG Inference Overhead.}}
\added{Since \cref{fig:infer_efficiency} benchmarks all methods without CFG, we measure the additional cost of enabling Hybrid CFG in \cref{tab:cfg_efficiency_h100}.}
\added{Enabling Hybrid CFG increases latency (\emph{Lat.}) by only $4.9$--$9.1\%$ and peak memory (\emph{Mem.}) by $0.01$\,GB across the evaluated lengths.}
%
\added{We attribute this modest latency increase to batched guidance exploiting spare GPU capacity, as the small motion-generation workload does not saturate the GPU.}
\added{Together with \cref{tab:cfg_main_results}, these results show that Hybrid CFG improves quality with modest latency overhead and nearly unchanged memory scaling.}
\section{Conclusion}
\label{sec:conclusion}
%
%
We present PASPA for long-horizon co-speech motion generation based on the observation that past history and the current segment demand different treatments.
PASPA realizes this via a hybrid design, summarizing past history into a persistent state while modeling the current segment with bidirectional attention.
Built on blockwise flow matching, PASPA incorporates multi-block supervision for effective training and a prefill-decode protocol with hybrid classifier-free guidance for stable, expressive inference.
%
%
Extensive public-benchmark evaluations show that PASPA achieves state-of-the-art FGD performance, preserving style and preventing temporal drift during extended motion generation. 

%

\begin{addedblock}
\section*{Acknowledgments}
This work was supported by the Alibaba Research Intern Program.
\end{addedblock}
{
    \small
    \bibliographystyle{ieeenat_fullname}
    \bibliography{neurips}
}

\clearpage
\appendix

\section{Empirical Motivation for the Asymmetry Between Past and Present}
\label{app:evidence_for_asymmetry}
\paragraph{Diagnostic Setup.}
As illustrated in \cref{fig:acf}, we employ autocorrelation analysis to investigate whether stylistic features and frame-level local dynamics exhibit distinct temporal persistence in co-speech motion generation.
The analysis is conducted on the training subset of BEAT2~\cite{liu2024emage}.

For each sequence, we convert SMPL-X rotations~\cite{loper2015smpl,pavlakos2019expressive} to 3D joint positions by forward kinematics.
%
To factor out orientation and scale variations, we respectively remove pelvic global yaw and normalize seven root-relative upper-body joints (neck, shoulders, elbows, wrists) by shoulder width. 
We use
\[
    \bfP_t =
    \left[
    \bfp_t^{\mathrm{neck}},
    \bfp_t^{\mathrm{left\text{-}shoulder}},
    \bfp_t^{\mathrm{right\text{-}shoulder}},
    \bfp_t^{\mathrm{left\text{-}elbow}},
    \bfp_t^{\mathrm{right\text{-}elbow}},
    \bfp_t^{\mathrm{left\text{-}wrist}},
    \bfp_t^{\mathrm{right\text{-}wrist}}
    \right]
    \in \mathbb{R}^{7 \times 3}
\]
to denote the normalized upper-body pose at frame $t$, where each $\bfp_t^{(\cdot)} \in \mathbb{R}^{3}$ is a root-local 3D joint position.
Both the style and local feature families are computed over the same trailing window $\mathcal{W}_t = \{t-W+1,\ldots,t\}$ with $W = 30$ frames (\ie, 1\,s).

The style feature $\bfs_t \in \mathbb{R}^{5}$ consists of five sub-features, defined as follows:
%
\begin{equation}
    \bfs_t =
    \left[
    s_t^{\mathrm{ext}},
    s_t^{\mathrm{span}},
    s_t^{\mathrm{height}},
    s_t^{\mathrm{lat}},
    s_t^{\mathrm{occ}}
    \right],
\end{equation}
whose five components are computed as follows:
\begin{itemize}
    \item \emph{Extension} measures the average distance from the elbows and wrists to the neck: 
    \begin{equation}
        s_t^{\mathrm{ext}}
        =
        \frac{1}{4W}
        \sum_{r\in\mathcal{W}_t}
        \sum_{a\in\mathcal{J}_{\mathrm{ext}}}
        \left\|
            \bfp_r^{a}
            -
            \bfp_r^{\mathrm{neck}}
        \right\|_2,
     \end{equation}
    with $
        \mathcal{J}_{\mathrm{ext}}
        =
        \{
        \mathrm{left\text{-}elbow},
        \mathrm{right\text{-}elbow},
        \mathrm{left\text{-}wrist},
        \mathrm{right\text{-}wrist}
        \}
    $.

    \item \emph{Span} measures the average distance between the two wrists:
    \begin{equation}
        s_t^{\mathrm{span}}
        =
        \frac{1}{W}
        \sum_{r\in\mathcal{W}_t}
        \left\|
            \bfp_r^{\mathrm{left\text{-}wrist}}
            -
            \bfp_r^{\mathrm{right\text{-}wrist}}
        \right\|_2 .
    \end{equation}

    \item \emph{Height} measures the average wrist height relative to the neck along the vertical axis:
    \begin{equation}
        s_t^{\mathrm{height}}
        =
        \frac{1}{2W}
        \sum_{r\in\mathcal{W}_t}
        \left[
            \left(
                \bfp_r^{\mathrm{left\text{-}wrist}}
                -
                \bfp_r^{\mathrm{neck}}
            \right)_y
            +
            \left(
                \bfp_r^{\mathrm{right\text{-}wrist}}
                -
                \bfp_r^{\mathrm{neck}}
            \right)_y
        \right].
    \end{equation}

    \item \emph{Laterality} measures the left-right asymmetry of hand extension:
    \begin{equation}
        s_t^{\mathrm{lat}}
        =
        \frac{1}{W}
        \sum_{r\in\mathcal{W}_t}
        \left(
        \left\|
            \bfp_r^{\mathrm{right\text{-}wrist}}
            -
            \bfp_r^{\mathrm{neck}}
        \right\|_2
        -
        \left\|
            \bfp_r^{\mathrm{left\text{-}wrist}}
            -
            \bfp_r^{\mathrm{neck}}
        \right\|_2
        \right).
    \end{equation}

    \item \emph{Occupancy} measures the spatial extent explored by the wrists within the window:
    \begin{equation}
        s_t^{\mathrm{occ}}
        =
        \log
        \left(
        \operatorname{tr}
        \left[
        \operatorname{Cov}
        \left(
            \left\{
            \bfp_r^{\mathrm{left\text{-}wrist}},
            \bfp_r^{\mathrm{right\text{-}wrist}}
            \right\}_{r\in\mathcal{W}_t}
        \right)
        \right]
        +
        \epsilon
        \right).
    \end{equation}
\end{itemize}
These components provide an interpretable diagnostic of stylistic speaker characteristics, whose temporal persistence can be compared with local motion dynamics.

The local motion feature $\bfl_t \in \mathbb{R}^{21}$ is defined as a window-level high-pass pose residual:
\begin{equation}
    \bfl_t
    =
    \operatorname{vec}
    \left(
        \bfP_t
        -
        \frac{1}{W}
        \sum_{r \in \mathcal{W}_t}
        \bfP_r
    \right)
    \in \mathbb{R}^{21}.
\end{equation}
Here, $\operatorname{vec}(\cdot)$ flattens the $7 \times 3$ joint-position matrix into a 21D vector.
This residual removes the local window mean and emphasizes fast-changing pose variations around the recent motion tendency.
By construction, $\bfl_t$ and $\bfs_t$ forms complementary frequency components over the same window $\mathcal{W}_t$, attributing any difference to the data rather than the operator.

For the alternative features shown in the inset of \cref{fig:acf}, with the local feature unchanged, we replace the hand-designed style feature $\bfs_t$ with the windowed raw pose mean:
\begin{equation}
    \bfs_t^{\mathrm{alt}}
    =
    \operatorname{vec}
    \left(
        \frac{1}{W}
        \sum_{r \in \mathcal{W}_t}
        \bfP_r
    \right)
    \in \mathbb{R}^{21},
\end{equation}
which is used to examine robustness against the choice of our hand-crafted style descriptors.

\paragraph{Feature Preprocessing.}
Before computing autocorrelation, we apply two preprocessing steps. 

%
First, to prevent speech energy from inflating the autocorrelation of features, we partial out its linear contribution from each feature dimension.
%
Let $\bfy_t \in \mathbb{R}^{D}$ denote any of $\bfs_t$, $\bfl_t$, or $\bfs_t^{\mathrm{alt}}$, and let $\rho_t$ denote the frame-aligned log-RMS speech energy (\ie, the natural logarithm of the per-frame root mean square of the waveform).
%
%
For each frame $t$, we build a lagged audio-energy vector:
\begin{equation}
    \bfu_t^{\mathrm{rms}}
    =
    \left[
    \rho_{t-L}, \ldots, \rho_t, \ldots, \rho_{t+L}
    \right]^{\intercal},
    \qquad L=15,
    \label{eq:acf_audio_lag}
\end{equation}
which covers a temporal range of $\pm 0.5$ seconds at 30 FPS.
For each feature dimension $d$, we fit a ridge regressor~\cite{hoerl1970ridge} from the lagged audio energy to the feature value:
\begin{equation}
    \hat{\boldsymbol{\beta}}_d
    =
    \arg\min_{\boldsymbol{\beta}}
    \sum_{t \in \mathcal{I}}
    {\lb
        \bfy_{t,d}
        -
        \lb\bfu_t^{\mathrm{rms}}\rb^{\intercal}
        \boldsymbol{\beta}
    \rb}^2
    +
    \lambda
    {\left\|
        \boldsymbol{\beta}
    \right\|}_2^2,
    \label{eq:acf_audio_ridge}
\end{equation}
where $\mathcal{I}$ denotes the valid frame indices with complete audio lags and we use $\lambda=1$.
We replace each feature dimension with its residual to obtain features without linear speech-energy contribution:
\begin{equation}
    \tilde{\bfy}_{t,d}
    =
    \bfy_{t,d}
    -
    \left(\bfu_t^{\mathrm{rms}}\right)^{\intercal}
    \hat{\boldsymbol{\beta}}_d .
    \label{eq:acf_audio_residual}
\end{equation}

Second, each feature dimension is standardized to zero mean and unit variance using per-speaker statistics, so that no single dimension dominates the autocorrelation.
For notational simplicity, we use $\bfy_t$ to denote the preprocessed feature sequence in the following.

\paragraph{Trace Autocorrelation.}  
Given a preprocessed multivariate feature sequence $\bfy_t \in \mathbb{R}^{D}$, where $\bfy_t$ can be either the style feature $\bfs_t$ or the local feature $\bfl_t$, we measure its temporal persistence using trace autocorrelation (ACF).
Let $\bar{\bfy}$ denote the temporal mean.
For a temporal lag $\tau$, the trace ACF is:
\begin{equation}
    r_{\bfy}(\tau)
    =
    \frac{
        \operatorname{tr}
        \lb
        \frac{1}{T-\tau}
        \sum_{t=1}^{T-\tau}
        \lb\bfy_t-\bar{\bfy}\rb
        {\lb\bfy_{t+\tau}-\bar{\bfy}\rb}^{\intercal}
        \rb
    }{
        \operatorname{tr}
        \lb
        \frac{1}{T}
        \sum_{t=1}^{T}
        \lb\bfy_t-\bar{\bfy}\rb
        {\lb\bfy_t-\bar{\bfy}\rb}^{\intercal}
        \rb
    }.
    \label{eq:trace_acf}
\end{equation}
This yields a variance-weighted ACF over feature dimensions, without selecting a coordinate.

After computing sequence-level ACFs, we aggregate these ACFs $r_{\bfy}(\tau)$ in two stages:
\begin{itemize}
    \item \emph{Within each speaker}, we average the per-sequence ACFs weighted by $\max(T_i - \tau, 0)$, where $T_i$ is the length of sequence $i$, so that short sequences contribute less at large lags.
    \item \emph{Across speakers}, we aggregate via the Fisher-$z$ transform~\cite{fisher1915frequency}:
    \begin{equation}
        \bar{r}_{\bfy}(\tau)
        =
        \tanh
        \lb
            \frac{1}{S}
            \sum_{s=1}^{S}
            \operatorname{arctanh}
            \lb
                r_{\bfy}^{s}(\tau)
            \rb
        \rb,
        \quad \tau > 0,
        \label{eq:fisher_z_acf}
    \end{equation}
    where $r_{\bfy}^{s}(\tau)$ is the within-speaker ACF of speaker $s$, and $S$ is the number of speakers.
    Specifically, we set $\bar{r}_{\bfy}(0)=1$ by definition.
\end{itemize}
Finally, we estimate 95\% confidence intervals by bootstrap resampling~\cite{efron1992bootstrap} over speakers (2000 resamples), which corresponds to the shaded regions shown in \cref{fig:acf}.
The same procedure is applied to both main and alternative features.

\paragraph{Observations and Implications.} 
\Cref{fig:acf} reveals a clear temporal-persistence asymmetry between the two feature types.
%
The style ACF $\bar{r}_{\bfs}(\tau)$ remains substantially positive beyond 100 frames.
Conversely, the local ACF $\bar{r}_{\bfl}(\tau)$ decays to near zero within a few tens of frames, indicating that frame-level pose dynamics are inherently short-term.
%
The pattern persists when replacing hand-crafted style descriptors with the windowed raw pose mean (inset), confirming the asymmetry is data-intrinsic rather than descriptor-induced.
%
This asymmetry motivates PASPA's core design: past history mainly carries slowly varying stylistic tendencies that should be summarized into a persistent state, whereas the current block contains fast-changing local dynamics that benefit from dense intra-block attention.

\section{Background: Chunkwise Parallel Form of Gated DeltaNet}
\label{app:detailed_parallel_form}



We denote the query, key, and value matrices as $\bfQ, \bfK \in \bbR^{T \times d_k}$ and $\bfV \in \bbR^{T \times d_v}$, where $T$ is the sequence length and $d_k, d_v$ denote the per-head query/key and value dimensions, respectively.
%

The chunkwise parallel form partitions $\bfQ, \bfK, \bfV$ into $N = T/K$ non-overlapping chunks of size $K$ (assuming $T$ is divisible by $K$). 
The $n$-th chunk ($n \in \{1, 2, \ldots, N\}$)  of $\bfQ$ is written as $\bfQ_{[n]}:=\bfq_{(n-1)K+1:nK+1}$ and analogously for $\bfK_{[n]}, \bfV_{[n]}$.
We further denote the $r$-th token within chunk $n$ as $\bfq^r_{[n]} := \bfq_{(n-1)K+r} \in \bbR^{d_k}$, with $\bfk^r_{[n]}, \bfv^r_{[n]}$ defined analogously. The scalar gating $\alpha^r_{[n]}$ and write-strength $\beta^r_{[n]}$ follow the same within-chunk indexing.
Finally, we define the cumulative gating product $\gamma^r_{[n]} := \prod_{j=1}^{r} \alpha^j_{[n]}$ for the first $r$ tokens of chunk $n$, with $\gamma^0_{[n]} := 1$ by convention.

The chunkwise parallel form computes the state update and output read for an entire chunk via: 
\begin{align}
    \bfS_{[n+1]} &= \overrightarrow{\bfS_{[n]}} + \Big(\widetilde{\bfU}_{[n]} - \overleftarrow{{\bfW}_{[n]}}\bfS^{\intercal}_{[n]}\Big)^{\intercal} \overrightarrow{\bfK_{[n]}}, \label{eq:gdn_state_parallel} \\
    \bfO_{[n]} &= \underbrace{\overleftarrow{\bfQ_{[n]}}\bfS^{\intercal}_{[n]}}_{\text{inter-chunk term}} + \underbrace{\big(\bfQ_{[n]}\bfK^{\intercal}_{[n]} \odot \bfM\big)\Big(\widetilde{\bfU}_{[n]} - \overleftarrow{{\bfW}_{[n]}}\bfS^{\intercal}_{[n]}\Big)}_{\text{intra-chunk term}}, \label{eq:gdn_output_parallel}
\end{align}
where $\bfM \in \bbR^{K \times K}$ is the intra-chunk causal mask with $\bfM_{ij} = 1$ if $i \geq j$ and $0$ otherwise. 
The output $\bfO_{[n]} \in \bbR^{K \times d_v}$ decomposes into two terms: an \emph{inter-chunk term} that reads from the block-start state $\bfS_{[n]}$ carrying information from previous chunks, and an \emph{intra-chunk term} capturing token-level interactions within the current chunk.
We detailed the core notations of \cref{eq:gdn_state_parallel,eq:gdn_output_parallel} below.

\paragraph{Arrow Notation.}
The left and right arrows denote cumulative gating coefficients within a chunk:
\begin{equation}
    \overleftarrow{\bfq^r_{[n]}} = \gamma^r_{[n]}\bfq^r_{[n]}, \quad
    \overleftarrow{{\bfw}^r_{[n]}} = \gamma^r_{[n]}{\bfw}^r_{[n]}, \quad
    \overrightarrow{\bfk^r_{[n]}} = \frac{\gamma^K_{[n]}}{\gamma^r_{[n]}}\bfk^r_{[n]}, \quad
    \overrightarrow{\bfS_{[n]}} = \gamma^K_{[n]}\bfS_{[n]},
\end{equation}
with the first three equations extending to stacked matrix forms $\overleftarrow{\bfQ_{[n]}}, \overleftarrow{\bfW_{[n]}}, \overrightarrow{\bfK_{[n]}} \in \bbR^{K \times d_k}$, whose $r$-th rows are $\overleftarrow{\bfq^r_{[n]}}, \overleftarrow{\bfw^r_{[n]}}, \overrightarrow{\bfk^r_{[n]}}$, respectively.
Intuitively, left arrows rescale each token by the cumulative gating from the start, while right arrows apply the remaining gating toward final position $K$.

\paragraph{WY coefficients and the UT transform.}
The matrices $\widetilde{\bfU}_{[n]} \in \bbR^{K \times d_v}$ and ${\bfW}_{[n]} \in \bbR^{K \times d_k}$ are the WY representation~\cite{bischof1987wy} coefficients for the cumulative generalized Householder product $\prod_{i}(\bfI - \beta^i_{[n]}\bfk^i_{[n]}\bfk^{i\intercal}_{[n]})$ that arises when unrolling \cref{eq:gdn_write} across a chunk. 
The matrices $\widetilde{\bfU}_{[n]}$ and $\bfW_{[n]}$ admit closed-form expressions via the UT transform~\cite{joffrain2006accumulating}, as follows:
\begin{align}
    \bfT_{[n]} &= \Big[\bfI + \text{strictLower}\big(\text{diag}(\beta_{[n]})(\bfGamma_{[n]} \odot \bfK_{[n]}\bfK^{\intercal}_{[n]})\big)\Big]^{-1} \text{diag}(\beta_{[n]}), 
    \label{eq:gdn_ut} \\
    {\bfW}_{[n]} &= \bfT_{[n]}\bfK_{[n]}, \qquad \widetilde{\bfU}_{[n]} = \bfT_{[n]}\bfV_{[n]}.  
    \label{eq:wu_matrix}
\end{align}
Here $\bfGamma_{[n]} \in \bbR^{K \times K}$ is the decay-aware mask with $(\bfGamma_{[n]})_{ij} = \frac{\gamma^i_{[n]}} {\gamma^j_{[n]}}$ for $i \geq j$ and $0$ otherwise, and $\text{strictLower}(\cdot)$ extracts the strictly lower-triangular part of a matrix, zeroing out other parts.

\paragraph{Remarks.}
\Cref{eq:gdn_state_parallel,eq:gdn_output_parallel}) consist of dense matrix multiplications and element-wise operations, making the chunkwise form well-suited for tensor-core acceleration. 
Flash Linear Attention (FLA)~\cite{yang2024fla} further provides Triton-accelerated implementations of such chunkwise kernels.




\section{Detailed Formulations of Loss Functions} 
\label{app:loss_details} 

In this section, we provide the exact mathematical definitions of the individual loss terms. 

\subsection{Details of Sequence-Level Losses}
The rotation loss $\mathcal{L}_{\mathrm{rot}}^n$ is the per-frame Mean Squared Error (MSE) between predicted and ground-truth motion representations within block $n$:
\begin{equation}
    \mathcal{L}_{\mathrm{rot}}^n
    = 
    \frac{1}{K}
    \sum_{i=1}^{K}
    \left\|
        \hat{\bfx}_0^{n,i}
        -
        \bfx_0^{n,i}
    \right\|_2^2 ,
    \label{eq:app_l_rot}
\end{equation}
where $\hat{\bfx}_0^{n,i}$ and $\bfx_0^{n,i}$ denote the predicted and ground-truth 6D motion representations with the global translation at the $i$-th frame of block $n$, respectively.


Following GlobalDiff~\cite{zhang2026mitigating}, we apply forward kinematics (FK) to map the rotations~\cite{zhou2019continuity} into 3D joint positions in the global space. 
We denote the predicted and ground truth FK joint positions at the $i$-th frame of block $n$ as $\hat{\bfmJ}_0^{n,i}$ and $\bfmJ_0^{n,i}$, respectively, and define two FK-based losses on top of them.

The joint loss $\mathcal{L}_{\mathrm{joint}}^n$ directly supervises absolute joint positions:
\begin{equation}
    \mathcal{L}_{\mathrm{joint}}^n
    =
    \frac{1}{K}
    \sum_{i=1}^{K}
    \left\|
        \hat{\bfmJ}_0^{n,i}
        -
        \bfmJ_0^{n,i}
    \right\|_2^2 .
    \label{eq:app_l_joint}
\end{equation}

The structural loss $\mathcal{L}_{\mathrm{struct}}^n$ instead supervises bone-level relations, enforcing consistency of pairwise bone-angle similarities to discourage unnatural bone stretching. 
Let $\mathcal{E}$ denote the set of bones in the kinematic tree, where each bone $e = (p(j), j)$ connects a child joint $j$ to its parent $p(j)$. The normalized predicted bone direction $\hat{\bfd}^{n,i}_{e}$ and ground truth bone direction ${\bfd}^{n,i}_{e}$ are
\begin{equation}
\hat{\bfd}^{n,i}_{e}
=
\frac{\hat{{\bfmJ}}^{n,i}_{j}-\hat{{\bfmJ}}^{n,i}_{p(j)}}
{\left\|\hat{{\bfmJ}}^{n,i}_{j}-\hat{{\bfmJ}}^{n,i}_{p(j)}\right\|_2+\epsilon},
\qquad
\bfd^{n,i}_{e}
=
\frac{{\bfmJ}^{n,i}_{j}-{\bfmJ}^{n,i}_{p(j)}}
{\left\|{\bfmJ}^{n,i}_{j}-{\bfmJ}^{n,i}_{p(j)}\right\|_2+\epsilon}.
\end{equation}
The predicted and ground truth pairwise bone-angle similarity matrix $\hat{\bfmA}^{n,i}$ and ${\bfmA}^{n,i}$ are 
\begin{equation}
\hat{\bfmA}^{n,i}_{e,e'}
=
\left(\hat{\bfd}^{n,i}_{e}\right)^\top
\hat{{\bfd}}^{n,i}_{e'},
\qquad
{\bfmA}^{n,i}_{e,e'}
=
\left({\bfd}^{n,i}_{e}\right)^\top
{\bfd}^{n,i}_{e'}.
\end{equation}
The structural loss is then defined as:
\begin{equation}
\mathcal{L}^{n}_{\mathrm{struct}}
=
\frac{1}{K|\mathcal{E}|^2}
\sum_{i=1}^{K}
\sum_{e,e'\in\mathcal{E}}
\rho\!\left(
\hat{{\bfmA}}^{n,i}_{e,e'}-{\bfmA}^{n,i}_{e,e'}
\right),
\end{equation}
where we use $\rho(\Delta) = |\Delta|$ (the original GlobalDiff paper uses the squared form $\rho(\Delta) = \Delta^2$).

\subsection{Details of Boundary Losses}
\label{app:boundary_loss}
\paragraph{Boundary Windows.}
The transition window $\hat{\bfb}_n$ at the boundary of the  $n$-th predicted block $\hat\bfx^n_0$ concatenates the last $W$ predicted frames of block $n$ with the first $W$ predicted frames of block $n+1$: 
\begin{equation}
    \hat{\bfb}_n
    := 
    \big( \hat{\bfb}_n^{(1)}, \ldots, \hat{\bfb}_n^{(2W)} \big),
    \quad \text{where} \quad
    \hat{\bfb}_n^{(j)}
    =
    \begin{cases}
        \hat{\bfx}_0^{n,\, K-W+j} & j = 1, \ldots, W, \\
        \hat{\bfx}_0^{n+1,\, j-W} & j = W+1, \ldots, 2W.
    \end{cases}
    \label{eq:bfm_bd_loss_slice}
\end{equation}
and $K$ is the block size. 
The ground truth window $\bfb_n$ is constructed analogously from the clean $n$-th block $\bfx_0^n$.
In all of our experiments, we set $W=4$.

\paragraph{Boundary Loss Formulations.}
Since adjacent blocks are denoised without temporal overlap during the training process, kinematic discontinuities may appear at the transition points. 
To address this, we impose the boundary loss on the concatenated $2W$-frame boundary window ($\hat{\bfb}_n, \bfb_n$) in \cref{eq:bfm_bd_loss_slice}.

Specifically, the boundary rotation loss $\mathcal{L}_{\mathrm{rot}}(\hat{\bfb}_n, \bfb_n)$ applies the MSE in \cref{eq:app_l_rot} to the $2W$ frames of the transition window. 
To explicitly guarantee temporal smoothness, we compute the first-order (velocity) and second-order (acceleration) finite differences on the FK-derived joint positions of the boundary window. 
The velocity matching loss $\mathcal{L}_{\mathrm{vel}}$ and acceleration matching loss $\mathcal{L}_{\mathrm{acc}}$ are:
\begin{align}
    \mathcal{L}_{\mathrm{vel}}(\hat{\bfb}_n, \bfb_n)
    &=
    \frac{1}{2W-1}
    \sum_{i=1}^{2W-1}
    \left\|
        \left(\hat{\bfmJ}_{\bfb_n}^{i+1} - \hat{\bfmJ}_{\bfb_n}^{i}\right)
        -
        \left(\bfmJ_{\bfb_n}^{i+1} - \bfmJ_{\bfb_n}^{i}\right)
    \right\|_2^2 ,
    \label{eq:app_l_vel} \\
    \mathcal{L}_{\mathrm{acc}}(\hat{\bfb}_n, \bfb_n)
    &=
    \frac{1}{2W-2}
    \sum_{i=1}^{2W-2}
    \left\|
        \left(\hat{\bfmJ}_{\bfb_n}^{i+2} - 2\hat{\bfmJ}_{\bfb_n}^{i+1} + \hat{\bfmJ}_{\bfb_n}^{i}\right)
        -
        \left(\bfmJ_{\bfb_n}^{i+2} - 2\bfmJ_{\bfb_n}^{i+1} + \bfmJ_{\bfb_n}^{i}\right)
    \right\|_2^2 ,
    \label{eq:app_l_acc}
\end{align}
where $\hat{\bfmJ}_{\bfb_n}^{i}$ and ${\bfmJ}_{\bfb_n}^{i}$ represents the FK joint position at the $i$-th frame within the boundary slice $\hat{\bfb}_n$ and $\bfb_n$, respectively. 
These kinematic terms collectively ensure that the generated motion remains physically plausible and temporally continuous across block transitions.

\section{Dataset Information}
\label{app:dataset_information}

\paragraph{\added{BEAT2.}} 
BEAT2, proposed by EMAGE~\cite{liu2024emage}, is a public mesh-level, motion-captured holistic co-speech motion benchmark derived from the earlier BEAT dataset~\cite{liu2022beat}, providing synchronized 3D body and facial motion paired with speech audio.
It contains about 60 hours of 3D motion data paired with speech from 25 speakers (12 female, 13 male), comprising 1{,}762 dialogue sequences that each last around 65 seconds on average.
%
Data representation includes SMPL-X~\cite{pavlakos2019expressive} body parameters (55 axis-angle joints, 3D translation, 4D foot-contact) and FLAME~\cite{li2017learning} facial expressions (100D coefficients).
Motion and speech are released at 30\,FPS and 16\,kHz respectively.
Following prior work~\cite{liu2024emage,mughal2025retrieving, zhang2026mitigating,zhang2025semtalk}, we adopt the same data partition and report results under two protocols: 
(1) The \textbf{1 Speaker} protocol trains and evaluates on the ``Scott'' subset, isolating style modeling within a single individual and enabling direct comparison with prior works that report single-speaker performance; 
(2) The \textbf{All Speakers} protocol uses the full speaker set jointly, probing cross-identity generalization.

\paragraph{\added{SHOW.}}
\added{SHOW~\cite{yi2023generating} is used as an additional benchmark to complement our primary evaluation on BEAT2.}
\added{It contains about 26.9 hours of in-the-wild talking videos from four speakers with diverse speaking styles.}
\added{Unlike the motion-captured annotations in BEAT2, SHOW provides pseudo-ground-truth motion reconstructed from monocular videos, represented as SMPL-X parameters covering body poses, hand gestures, and facial expressions.}
\added{Motion is provided at 30\,FPS with synchronized speech audio.}
\added{We follow the gesture-evaluation protocol of DiffSHEG~\cite{chen2024diffsheg} for the SHOW experiments.}

\begin{table}[htbp]
    \centering
    \small 
    \newcolumntype{L}{>{\raggedright\arraybackslash}X}
    \newcolumntype{Y}{>{\centering\arraybackslash}X}
    \renewcommand{\arraystretch}{1.1}
    \setlength{\tabcolsep}{12pt} 
    
    \caption{
        Implementation details of our best training and inference configurations on BEAT2.
    }
    \label{tab:training_details}
    
     \begin{tabularx}{\linewidth}{@{}llL@{}}
        \toprule
        \textbf{Category} & \textbf{Hyperparameter} & \textbf{Setting / Value} \\
        \midrule
        
        \multirow{7}{*}{\textbf{Dataset \& Setup}}
        & Skeleton                   & 55 joints (30 hands, 25 torso) \\
        & Translation                & 3D global root translation \\
        & Expression                 & 100-dim FLAME blendshapes \\
        & Rotation Representation    & axis-angle (raw) $\rightarrow$ rot6d~\cite{zhou2019continuity} \\
        & Motion Dimension           & $55 \times 6 + 3 = 333$ \\
        & Audio Features             & WavLM~\cite{chen2022wavlm}, 16\,kHz, 1024-dim, pre-extracted \\
        & \# Speakers                & 25 identities \\
        \midrule
        
        \multirow{2}{*}{\textbf{Hardware}} 
        & Devices \& Precision          & 4$\times$ NVIDIA A800 (40 GB) GPUs, Float32 \\
        & Parallelism            & Data parallel (DDP) \\
        \midrule

        \multirow{7}{*}{\textbf{Architecture}} 
        & Parameters / MACs          & 65.158 M, 16.633 G \\
        & Hidden Dim / Depth / Heads & $384$ / $6$ / $8$ \\
        & Time (and ID) Embedding    & 768 Channels\\
        & History Module             & Blockwise Gated DeltaNet \\
        & Self-Attention Module      & Blockwise Multi-Head Self-Attention \\
        & Norm / Positional Encoding & RMSNorm~\cite{zhang2019root} / RoPE~\cite{su2024roformer} \\
        & Block Size ($K$)           & 64 frames \\
        \midrule
        

        
        \multirow{10}{*}{\textbf{Training}}
        & Training Length    & $1{,}000$ epochs ($316{,}000$ steps) \\
        & Training Time      & $\sim$30 hours on 4 GPUs ($\sim$120 GPU-hours) \\
        & Checkpoint Selection   & {Every 10 epochs; best-FGD checkpoint reported} \\
        & Batch Size         & 72 per GPU (global: 288) \\
        & Training Sequence  & 256 frames (\ie, 4 contiguous blocks) \\
        & CFG Dropout        & cond $0.1$; hist: keep $0.7$ / null $0.1$ / interp.\ $0.2$ \\
        & Gradient Clipping  & max-norm $= 1.0$ \\
        & Adaptive Loss Weighting    & \ding{52} (\emph{Enabled}) \\
        & Time Sampling              & Weakly Monotonic Logit-normal~\cite{esser2024scaling} \\
        & Logit-Normal               & mean $\mu=-1.0$, std $\sigma=1.0$ \\
        \midrule

        \multirow{5}{*}{\textbf{Optimization}}
        & Optimizer / EMA       & Muon~\cite{liu2025muon,jordan2024muon} + AdamW~\cite{kingma2014adam} / \ding{55} (\emph{Disabled})\\
        & Learning Rate    & $10^{-3} \to 10^{-4}$ at step $5{,}000$ \\
        & Muon Settings    & momentum $=0.95$, RMS-adjusted LR \\
        & AdamW Settings   & $\beta_1=0.9$, $\beta_2=0.999$ \\
        & Weight Decay     & $0.01$ \\
        \midrule
        
        \multirow{3}{*}{\textbf{Inference}} 
        & Sampling Steps / Seed Length        & 10 / 8\\
        & CFG Mode / CFG Scales  & Hybrid (\emph{Interaction}) / $w_\mathrm{cond}=1.5, w_\mathrm{hist}=1.5$ \\
        & Activation Interval    & $t \in [0.1, 0.6]$ \\
       
        \bottomrule
    \end{tabularx}
\end{table}

\section{Implementation Details}
\label{app:implementation_details}

\subsection{Implementation Details of Main Results}
\label{app:imple_main_results}

\Cref{tab:training_details} summarizes \added{training and inference configurations on BEAT2, our primary benchmark}.
\added{We apply these settings to SHOW for supplementary evaluation of PASPA's effectiveness beyond BEAT2.}

\paragraph{Dataset and Motion Representation.}
We train and evaluate on BEAT2~\cite{liu2024emage}, comprising 25 speakers with synchronized speech, body motion, and expressions.
Each motion frame consists of $55$ skeleton joints (25 torso, 30 hands), a global root translation, and a 100-dimensional FLAME \added{expression vector}~\cite{li2017learning}.
%
Following common practice, native axis-angle rotations are converted to rot6d~\cite{zhou2019continuity}. 
\added{The rot6d rotations}~\cite{zhou2019continuity} \added{and global root translation form a 333-dimensional motion vector.}
%
%
\added{For SHOW~\cite{yi2023generating}, we expand the hand PCA coefficients into axis-angle rotations and align the poses with BEAT2's 55-joint ordering, followed by conversion to global rot6d.}
\added{Translation channels are set to zero, while the 100-dimensional expression coefficients are retained.}
\added{For both datasets,} speech is processed at $16$\,kHz, yielding $1024$-dim WavLM features~\cite{chen2022wavlm}.

\paragraph{Architecture and Hardware.}
We adopt a $6$-layer backbone shown in \cref{fig:network} with hidden dimension $384$ and $8$ attention heads.
It primarily consists of a PASPA module (\cref{sec:paspa_block}), with BGDN for history propagation and BMHSA for current-block modeling.
We use RMSNorm~\cite{zhang2019root} for normalization and RoPE~\cite{su2024roformer} for positional encoding within BMHSA (see \cref{sec:intra_block}).
The block size is set to $K=64$ by default, the empirically optimal value in \cref{tab:block_size_seq_len}.
\added{Speaker IDs are one-hot encoded and mapped to separate learned embeddings in the expression predictor and motion backbone. In the latter, the speaker and timestep embeddings are concatenated into a 768-dimensional conditioning vector.}

\added{For both datasets, a speaker-conditioned network predicts 100-dimensional expression coefficients from WavLM features.}
\added{On BEAT2, we reuse GlobalDiff's pretrained audio-to-expression network~\cite{zhang2026mitigating}.
On SHOW, we train a TalkSHOW-based face predictor~\cite{yi2023generating} using WavLM features to predict the expression.}
\added{WavLM and expression predictors remain frozen during PASPA training.}

\added{The BEAT2 configuration has} $65.158$\,M parameters and $16.633$\,G multiply-accumulate operations (MACs) per forward pass.
The MACs are measured via \texttt{ptflops} at batch size $1$ and sequence length $256$, with all Triton kernels replaced by their PyTorch fallbacks.
Training is distributed via data parallelism on 4 NVIDIA A800 GPUs using Float32 precision.

\paragraph{Training.}
%
We train for $1{,}000$ epochs \added{on both datasets ($316{,}000$ steps on BEAT2 and $82{,}000$ on SHOW)}.
\added{On BEAT2,} main experiments and $500$-epoch ablations and analyses take approximately $30$ and $15$ hours, respectively, on $4$ GPUs.
\added{Following the standard reporting protocol used by prior work~\cite{zhang2026mitigating,zhang2025semtalk},} we save checkpoints every $10$ epochs and report results from the checkpoint which achieves the best FGD \added{on each dataset}.
%
Each GPU processes $72$ sequences of $256$ frames (\ie, $4$ blocks), yielding a global batch size of $288$.
%
As detailed in \cref{sec:blockwise_flow_matching}, we sample timesteps from a logit-normal distribution~\cite{esser2024scaling} ($\mu = -1.0$, $\sigma = 1.0$) and enforce a weakly monotonic order across blocks, with both designs empirically validated in \cref{tab:ablation_train,tab:lognorm_sweep}.
The training loss uses adaptive loss weighting (see \cref{eq:bfm_total_loss}) to balance contributions across timesteps and blocks, which is proven crucial in \cref{tab:ablation_train}.
%
%
The condition (\texttt{cond}) is dropped with $0.1$ probability, while the history state (\texttt{hist}) is assigned three modes: kept (\texttt{keep}, $0.7$), nulled (\texttt{null}, $0.1$), or linearly interpolated to null (\texttt{interp}, $0.2$).
To avoid extreme gradient updates, we rescale gradients to unit L2 norm.

\paragraph{Optimization.}
We optimize PASPA via a mixed Muon-AdamW optimizer~\cite{liu2025muon,jordan2024muon,kingma2014adam} for fast convergence, omitting weight EMA (exponential moving average).
Following standard practice, Muon~\cite{liu2025muon,jordan2024muon} optimizes the hidden matrix-shaped weights, while AdamW~\cite{kingma2014adam} handles the remaining parameters, including the input embeddings, the final output projection, and the $1$D gains and biases. 
%
We further place BGDN under AdamW, as we find Muon ineffective on it.
Muon uses momentum $0.95$ with RMS-matched learning-rate adjustment to keep the Muon update scale comparable to AdamW. 
AdamW uses $\beta_1 = 0.9$ and $\beta_2 = 0.999$. 
Both optimizers apply weight decay $0.01$.
The shared learning rate is $10^{-3}$, dropping to $10^{-4}$ after $5{,}000$ steps to accelerate early training.
%

\paragraph{Inference.}
We use 10 denoising steps.
Following common practice~\cite{liu2024emage,zhang2025semtalk,zhang2026mitigating},  we prefill an 8-frame seed motion (see line~1 of \cref{alg:rollout_cfg}) into the state to initialize the rollout. 
We adopt Hybrid CFG under the \emph{Interaction} mode (\ie, keeping second-order joint term in line~7 of \cref{alg:rollout_cfg}; concrete form in \cref{tab:cfg_decomp_formulation}).
We use guidance scales $w_\mathrm{cond} = 1.5$ and $w_\mathrm{hist} = 1.5$ for the condition and the history, validated in \cref{tab:cfg_scale_sweeps}.
%
Guidance is restricted to $t \in [0.1, 0.6]$, tuned via \cref{tab:cfg_interval}.

\subsection{Implementation Details of Other Experiments}
\label{app:imple_other_exp}

\paragraph{Long-rollout endpoint analysis.}
We conduct the long-rollout style preservation analysis in \cref{tab:long_rollout}.
This analysis requires sequences up to $60$\,s, and restricting the evaluation to a single speaker leaves too few such clips for stable statistics.
We therefore evaluate on a subset of the full test set comprising all sequences exceeding $60$\,s.
%
For each rollout horizon $T \in \{10,30,60\}$\,s, metrics are computed on the terminal $10$\,s window, corresponding to $[0,10]$\,s, $[20,30]$\,s, and $[50,60]$\,s respectively.
%
We use two baselines: \emph{Short Seed} (corresponding to Short-Seed Truncation in \cref{fig:teaser}, which keeps only the last $8$ frames as history) and \emph{Token Attention} (corresponding to Long-Context Token Attention in \cref{fig:teaser}, which retains all past frames as tokens for attention).
%
%
\Cref{fig:app_long_rollout_curves} further reports rollout trends with additional metrics, using $95\%$ confidence intervals from $1{,}000$ clip-level bootstrap resamples.

\paragraph{Inference Efficiency Benchmark.}  
We evaluate scalability on a single NVIDIA 4090 GPU under Float32 precision in \cref{fig:infer_efficiency}, using the same configuration in \cref{tab:training_details}, with CFG disabled for all methods.
\emph{PASPA} refers to our default setting (using prefill-decode rollout and state caching), while \emph{PASPA w/o PD} disables prefill-decode separation and re-prefills the full history at every denoising step.
%
We also include two baselines from \cref{fig:teaser}: \emph{Token Attention} and \emph{Short Seed}.
We sweep rollout length over $\{10, 30, 60, 120, 300, 600\}\,$s ($300$ to $18{,}000$ frames at $30\,$fps).
%
For each configuration, after $10$ warm-ups, we record $50$ timed runs, measuring per-call latency via \texttt{time.perf\_counter} (with \texttt{torch.cuda.synchronize}) and peak memory via \texttt{torch.cuda.max\_memory\_allocated}.
\added{For the CFG-overhead analysis in \cref{tab:cfg_efficiency_h100}, we benchmark PASPA with and without Hybrid CFG on one NVIDIA H100 using an input batch size of $1$, following the same timing and memory protocol.}

\begin{figure}[t] 
    \centering
    \includegraphics[width=\linewidth]{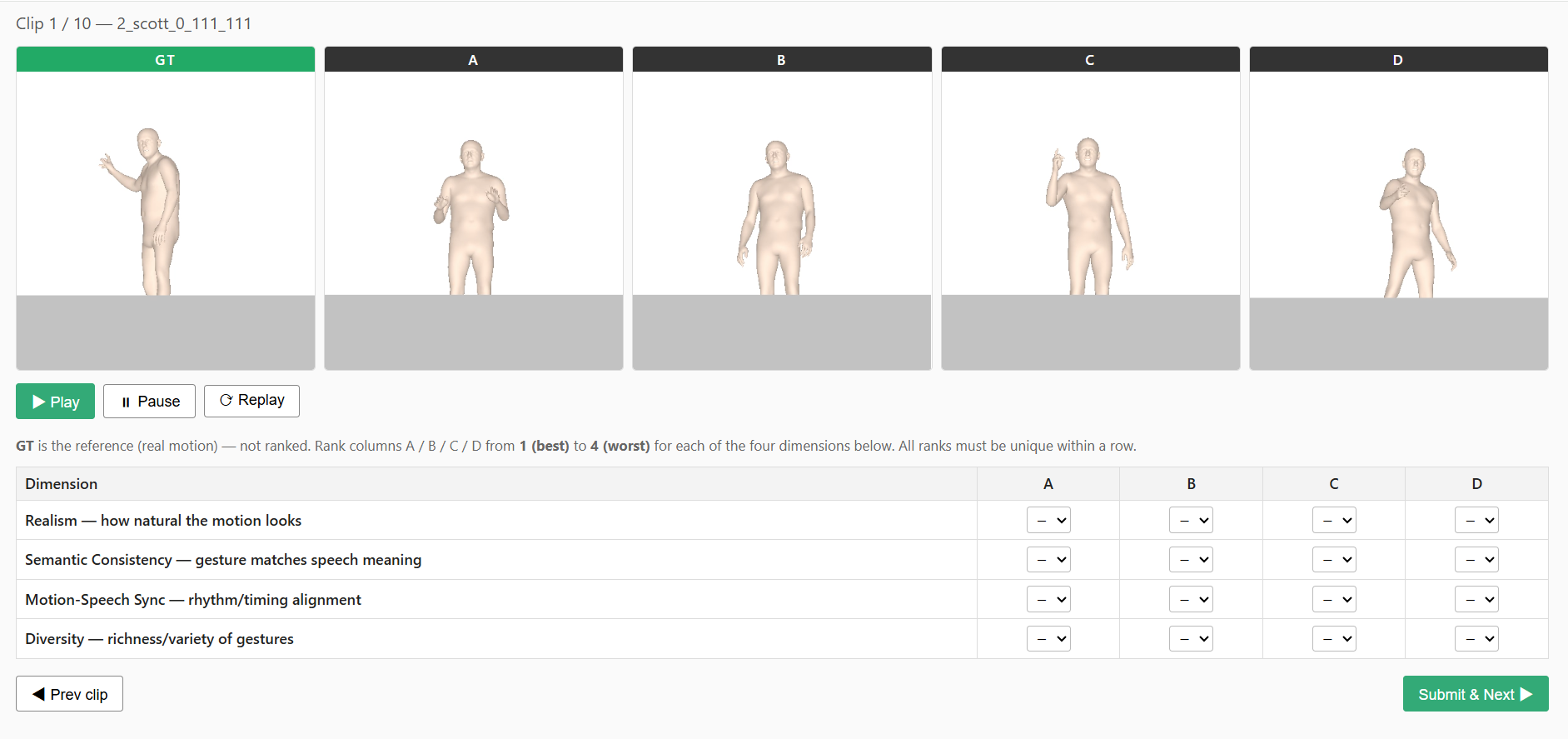}
    \caption{
        User study interface.
        Participants rank four anonymized methods (A/B/C/D, randomized per clip) produced by SemTalk, GestureLSM, EMAGE and PASPA, against a non-rankable ground-truth reference (GT), along four criteria: realism, semantic consistency, motion-speech sync, and diversity.
        \label{fig:user_study_page}
    }
    
\end{figure}


\subsection{Implementation Details of the User Study}
\label{app:imple_user_study}
We conduct a blind ranking study through a custom web interface (see \cref{fig:user_study_page}) to assess perceptual quality. 
A total of 18 participants take part in the study, yielding 167 valid ranking groups after filtering incomplete annotations.
%
For each clip, participants are shown the ground-truth motion together with four anonymized generated motions produced by SemTalk~\cite{zhang2025semtalk}, GestureLSM~\cite{liu2025gesturelsm}, EMAGE~\cite{liu2024emage}, and our PASPA, with method order randomized per clip.
Participants rank all four methods from best (1) to worst (4) along four criteria: 
(1) realism,
(2) semantic consistency with speech,
(3) motion-speech synchronization, 
and (4) motion diversity, 
with unique ranks enforced per criterion.
We report the mean rank in \cref{app:user_study} (lower is better), where PASPA achieves the best rank across all four criteria.

\section{Additional Experimental Results}
\label{app:additional_experimental_results}

\subsection{Additional Ablation Studies}
\label{app:additional_ablation_studies}
\begin{table*}[t] 
    \centering
    \renewcommand{\arraystretch}{1.25} 
    \newcolumntype{Y}{>{\centering\arraybackslash}X}
    \setlength{\tabcolsep}{4pt} 
\begin{minipage}[t]{0.52\linewidth} 
    \centering
    \caption{
        Ablation on training time distribution, including uniform and logit-normal sampling.
    }
    \label{tab:lognorm_sweep}
    \begin{tabularx}{\linewidth}{@{} l  *{4}{Y} @{}}
        \toprule
        \textbf{Time Sampling}
        & \textbf{FGD}\,$\downarrow$ & \textbf{BC}\,$\uparrow$
        & \textbf{Div}\,$\uparrow$ & \textbf{MSE}\,$\downarrow$ \\
        \midrule
        Uniform                          & 5.216 & 0.734 & 14.26 & 5.761 \\
        \hdashline
        \addlinespace
        \multicolumn{5}{@{}l}{\textit{Logit-Normal sweep on mean $\mu$ with std $\sigma=1.0$}} \\
        \quad $\mu = \phantom{-}0.0$     & 4.952 & 0.736 & 12.56 & 5.913 \\
        \quad $\mu = -0.2$               & 5.552 & \best{0.742} & 13.00 & 5.881 \\
        \quad $\mu = -0.4$               & 5.320 & 0.738 & 13.74 & 5.843 \\
        \quad $\mu = -0.6$               & 4.908 & 0.734 & 13.28 & 5.773 \\
        \quad $\mu = -0.8$               & 4.548 & 0.730 & 12.24 & 5.795 \\
        \rowcolor{graybg}
        \textbf{\quad $\mu = -1.0$}      & \best{4.450} & 0.739 & 12.36 & 5.763 \\
        \quad $\mu = -1.2$               & 5.128 & 0.738 & \best{14.35} & \best{5.660} \\
        \bottomrule
    \end{tabularx}
\end{minipage}
    \hfill 
\begin{minipage}[t]{0.46\linewidth}
    \centering
    \caption{
        Ablation on block size ($K$) and training window length ($T$) of our blockwise FM.
    }
    \label{tab:block_size_seq_len}
    \begin{tabularx}{\linewidth}{@{} l  *{4}{Y} @{}}
        \toprule
        \textbf{Variant}
        & \textbf{FGD}\,$\downarrow$ & \textbf{BC}\,$\uparrow$
        & \textbf{Div}\,$\uparrow$   & \textbf{MSE}\,$\downarrow$ \\
        \midrule
        \multicolumn{5}{@{}l}{\textit{Block size sweep with fixed $T=256$}} \\
        \quad $K=16$ & 5.030 & \best{0.744} & 13.58 & 9.158 \\
        \quad $K=32$ & 5.188 & 0.730 & \best{13.89} & 13.941 \\
        
        \rowcolor{graybg}
        \quad \textbf{$K=64$}
         & \best{4.450} & 0.739 & 12.36 & \best{5.763} \\
        
        \midrule
        \multicolumn{5}{@{}l}{\textit{Training window sweep with fixed $K=64$}} \\
        \quad $T=128$ & 4.955 & 0.732 & \best{14.70} &\best{5.660} \\
        
        \rowcolor{graybg}
        \quad \textbf{$T=256$}
        & \best{4.450} & \best{0.739} & 12.36 & 5.763 \\
         
        \quad $T=512$ & 5.811 & 0.731 & 14.42 & 5.807 \\
        \quad $T=768$ & 4.706 & 0.735 & 14.20 & 5.768 \\
        \bottomrule
    \end{tabularx}
\end{minipage}
\end{table*}

\subsubsection{Time Distribution}
\Cref{tab:lognorm_sweep} provides the parameter study supporting our LogitN choice (\ie, logit-normal time sampling~\cite{esser2024scaling}) in the main ablation (see \cref{tab:ablation_train}).
All rows share the same training recipe (\ie, Mono $+$ AdaW $+$ SG 
in \cref{tab:ablation_train}) and differ only in the time sampling distribution mentioned in \cref{sec:blockwise_flow_matching}.
The Uniform row corresponds to the $+$SG variant in \cref{tab:ablation_train}, while the LogitN rows fix the standard deviation $\sigma=1.0$ and vary the mean $\mu$.
Choosing a moderately negative LogitN mean improves FGD over uniform sampling, with $\mu=-1.0$ achieving the best FGD.
However, an overly negative mean (\eg, $\mu=-1.2$) worsens FGD, suggesting that time sampling should not be biased too aggressively.
Although other $\mu$ values are favored by individual secondary metrics (\eg, BC, Div and MSE), we use FGD as the primary selection criterion and therefore adopt $\mu=-1.0$.
\subsubsection{Block Size and Training Sequence Length}
\Cref{tab:block_size_seq_len} studies the sensitivity to the block size $K$ and the training window length $T$.
For the block-size sweep, we fix $T=256$ and vary $K$, while for the window-length sweep, we fix $K=64$ and vary $T$.
In the block-size sweep, $K=64$ achieves the best FGD, whereas smaller blocks degrade FGD and substantially increase MSE.
This suggests that overly short blocks fragment the local motion segment and limit the dense bidirectional interactions needed for modeling fast-changing intra-block dynamics.
In the window-length sweep, $T=256$ gives the best FGD.
A shorter window ($T=128$) remains competitive on some secondary metrics, but its worse FGD indicates that limited histories hurt the primary motion quality.
%
Conversely, increasing the window length (\eg, $T=512$) does not further improve FGD, which we attribute to the larger number of blocks in each training window, introducing more inter-block dependencies and making multi-block denoising harder to learn.
We therefore use $K=64$ and $T=256$ as the default setting.

\subsubsection{Classifier-Free Guidance}
\label{app:cfg_ablation}
\begin{table}[t]
    \centering
    \renewcommand{\arraystretch}{1.20}
    \newcolumntype{Y}{>{\centering\arraybackslash}X}
    
    \caption{
        CFG scale ablations for Condition CFG and Hybrid CFG.
        Cond CFG denotes condition CFG.
        Both panels use the same activation interval $t\in[0.1,0.6]$.
    }
    \label{tab:cfg_scale_sweeps}

    \begin{minipage}[t]{0.5\linewidth}
        \centering
        \vspace{8pt}
        \subcaption{Condition CFG}
        \label{tab:cond_cfg_scale}
        \begin{tabularx}{\linewidth}{@{}l YYYY@{}}
            \toprule
            $\boldsymbol{w_{\mathrm{cond}}}$
            & \textbf{FGD}\,$\downarrow$
            & \textbf{BC}\,$\uparrow$
            & \textbf{Div}\,$\uparrow$
            & \textbf{MSE}\,$\downarrow$ \\
            \midrule
            --  & 4.450        & 0.739        & 12.36        & \best{5.763} \\
            \hdashline
            \multicolumn{5}{@{}l}{\textit{Condition CFG guidance}} \\
            0.5 & 4.151        & 0.739        & 13.28        & 5.811 \\
            1.0 & 3.937        & 0.741        & 13.46        & 5.814 \\
            \rowcolor{graybg}
            1.5 & \best{3.883} & 0.740        & 13.39        & 5.823 \\
            2.0 & 4.060        & 0.741        & 13.19        & 5.803 \\
            2.5 & 4.235        & 0.742        & 13.14        & 5.825 \\
            3.0 & 4.080        & 0.748        & 13.27        & 5.811 \\
            3.5 & 4.208        & 0.746        & 13.50        & 5.827 \\
            4.5 & 4.640        & \best{0.753} & \best{13.66} & 5.810 \\
            \bottomrule
        \end{tabularx}
    \end{minipage}
    \hfill
    \begin{minipage}[t]{0.48\linewidth}
        \centering
        \subcaption{Hybrid CFG ablation}
        \label{tab:hybrid_cfg_scale}
        \vspace{-2pt}
        \begin{tabularx}{\linewidth}{@{}l YYYY@{}}
            \toprule
            {\small $\boldsymbol{(w_{\mathrm{cond}}, w_{\mathrm{hist}})}$}
            & \textbf{FGD}\,$\downarrow$
            & \textbf{BC}\,$\uparrow$
            & \textbf{Div}\,$\uparrow$
            & \textbf{MSE}\,$\downarrow$ \\
            \midrule
            $(1.5, \text{--})$ & 3.883        & 0.740        & 13.39        & 5.823 \\
            \hdashline
            \multicolumn{5}{@{}l}{\textit{Hist scale sweep ($w_{\mathrm{cond}}=1.5$)}} \\
            $(1.5, 0.5)$ & 4.163        & 0.741        & 13.17        & 5.788 \\
            $(1.5, 1.0)$ & 3.989        & 0.743        & 13.17        & 5.782 \\
            \rowcolor{graybg}
            $\mathbf{(1.5, 1.5)}$ & \best{3.532} & 0.745        & 13.30        & \best{5.779} \\
            $(1.5, 2.0)$ & 4.156        & 0.744        & \best{13.77}        & 5.797 \\
            $(1.5, 2.5)$ & 5.603        & \best{0.751} & 13.76        & 5.779 \\
            \hdashline
            \multicolumn{5}{@{}l}{\textit{Cond scale sweep ($w_{\mathrm{hist}}=1.5$)}} \\
            $(0.5, 1.5)$ & 3.884        & 0.739        & 13.28        & 5.803 \\
            $(1.0, 1.5)$ & 3.611        & 0.740        & 13.20        & 5.781 \\
            $(2.0, 1.5)$ & 3.657        & 0.745        & 13.57        & 5.784 \\
            $(2.5, 1.5)$ & 3.769        & 0.748        & {13.74} & 5.790 \\
            \bottomrule
        \end{tabularx}
    \end{minipage}
\end{table}

\paragraph{Conditional Classifier-Free Guidance.}
\Cref{tab:cond_cfg_scale} ablates Condition CFG.
The first row is trained without CFG-aware dropout and evaluated without inference-time guidance, while $w_{\mathrm{cond}}=1.0$ is trained with CFG-aware dropout but evaluated without inference-time guidance.
This comparison shows that condition dropout alone substantially improves FGD from 4.450 to 3.937.
We attribute this gain to condition dropout acting as a training-time data augmentation, which improves robustness to imperfect or weakened conditioning.
Increasing the scale to $w_{\mathrm{cond}}=1.5$ further improves FGD to 3.883, showing that moderate inference-time guidance provides an additional gain.
However, larger scales degrade FGD (motion quality), despite improving BC and Div, suggesting over-guidance.

\paragraph{Hybrid Classifier-Free Guidance.}
\Cref{tab:hybrid_cfg_scale} ablates Hybrid CFG with the two CFG scales $w_{\mathrm{cond}}$ and $w_{\mathrm{hist}}$.
The first row $(1.5, \text{--})$ is the Condition CFG optimum from \cref{tab:cond_cfg_scale}, which uses the checkpoint trained with condition (\ie, speech and expression) dropout only and serves as the baseline.
All subsequent rows use the checkpoint trained with both condition and history dropout mentioned in \cref{sec:inference}.
We first sweep the history scale with $w_{\mathrm{cond}}=1.5$ fixed.
Adding history dropout in training alone, without history guidance (\ie, $(1.5, 1.0)$), slightly degrades FGD from $3.883$ to $3.989$, indicating that the history dropout does not directly improve quality.
%
However, applying history guidance unlocks a substantial improvement, with $(1.5,1.5)$ achieving the best FGD (\ie, $3.532$).
Over-guidance is more severe along the history axis than in Condition CFG: $w_{\mathrm{hist}}=2.5$ substantially degrades FGD to $5.603$.
We then sweep $w_{\mathrm{cond}}$ with $w_{\mathrm{hist}}=1.5$ fixed:  $(1.5, 1.5)$ remains the FGD optimum, while other values of $w_{\mathrm{cond}}$ yield slightly higher FGD but can improve secondary metrics (\eg, Div).
We adopt $(w_{\mathrm{cond}}, w_{\mathrm{hist}}) = (1.5, 1.5)$ as the default. 

\Cref{tab:cfg_decomp_formulation} summarizes four ways to compose the condition- and history-guidance directions in Hybrid CFG, and \cref{tab:hybrid_cfg_mode} reports their performance.
\Cref{tab:cfg_decomp_formulation} adopts a simplified notation: $\hat{\bfx}_0$, $\hat{\bfx}_0^{c}$, $\hat{\bfx}_0^{h}$, and $\hat{\bfx}_0^{c,h}$ denote the unconditional, condition-only, history-only, and full predictions, respectively.
\emph{State First} and \emph{Cond First} apply the two guidance directions sequentially in opposite orders: \emph{State First} first moves from $\hat{\bfx}_0$ to $\hat{\bfx}_0^{h}$ and then toward $\hat{\bfx}_0^{c,h}$, while \emph{Cond First} reverses the order.
\emph{Independent} treats both guidance directions as parallel marginal contributions relative to $\hat{\bfx}_0$, while \emph{Interaction} additionally adds a second-order term capturing their joint effect.
As shown in \cref{tab:hybrid_cfg_mode}, \emph{Interaction} achieves the best FGD, supporting our choice in \cref{alg:rollout_cfg}.
We attribute this to the explicit modeling of the joint effect between condition and history beyond their marginal contributions. 

\begin{table}[t]
    \centering
    \renewcommand{\arraystretch}{1.25}
    \setlength{\tabcolsep}{5pt}
    \newcolumntype{L}{>{\raggedright\arraybackslash}X}
    \caption{
        Formulations of Hybrid CFG decomposition modes.
        $\hat{\bfx}_0$, $\hat{\bfx}_0^{c}$, $\hat{\bfx}_0^{h}$, and $\hat{\bfx}_0^{c,h}$
        denote predicted clean motions under neither condition nor history,
        condition only, history only, and both condition and history, respectively.
    }
    \label{tab:cfg_decomp_formulation}
    \begin{tabularx}{\linewidth}{@{}l L@{}}
        \toprule
        \textbf{Mode} & \textbf{Guided prediction} \\
        \midrule
        \emph{State First}
        &
        $
        \hat{\bfx}_0
        + w_{\mathrm{hist}}(\hat{\bfx}_0^{h}-\hat{\bfx}_0)
        + w_{\mathrm{cond}}(\hat{\bfx}_0^{c,h}-\hat{\bfx}_0^{h})
        $
        \\

        \emph{Cond First}
        &
        $
        \hat{\bfx}_0
        + w_{\mathrm{cond}}(\hat{\bfx}_0^{c}-\hat{\bfx}_0)
        + w_{\mathrm{hist}}(\hat{\bfx}_0^{c,h}-\hat{\bfx}_0^{c})
        $
        \\

        \emph{Independent}
        &
        $
        \hat{\bfx}_0
        + w_{\mathrm{cond}}(\hat{\bfx}_0^{c}-\hat{\bfx}_0)
        + w_{\mathrm{hist}}(\hat{\bfx}_0^{h}-\hat{\bfx}_0)
        $
        \\

        \emph{Interaction}
        &
        $
        \hat{\bfx}_0
        + w_{\mathrm{cond}}(\hat{\bfx}_0^{c}-\hat{\bfx}_0)
        + w_{\mathrm{hist}}(\hat{\bfx}_0^{h}-\hat{\bfx}_0)
        + w_{\mathrm{cond}}w_{\mathrm{hist}}
        (\hat{\bfx}_0^{c,h}-\hat{\bfx}_0^{c}-\hat{\bfx}_0^{h}+\hat{\bfx}_0)
        $
        \\
        \bottomrule
    \end{tabularx}
\end{table}
\begin{table}[t]
    \centering
    \renewcommand{\arraystretch}{1.20}
    \setlength{\tabcolsep}{3.5pt}
    \newcolumntype{Y}{>{\centering\arraybackslash}X}
    

            
            
            
    \begin{minipage}[t]{0.49\linewidth}
        \centering
        \caption{
            Ablation on Hybrid CFG decomposition.
            All variants use Hybrid CFG with $(w_{\mathrm{cond}}, w_{\mathrm{hist}})=(1.5,1.5)$ and the activation interval $t\in[0.1,0.6]$.
            \label{tab:hybrid_cfg_mode}
        }
        \begin{tabularx}{\linewidth}{@{}l YYYY@{}}
            \toprule
            \textbf{Mode}
            & \textbf{FGD}\,$\downarrow$
            & \textbf{BC}\,$\uparrow$
            & \textbf{Div}\,$\uparrow$
            & \textbf{MSE}\,$\downarrow$ \\
            \midrule
            \emph{Independent}
            & 3.967 & 0.742 & \best{13.76} & 5.800 \\
            
            \emph{Cond First}
            & 3.704 & 0.743 & 13.35 & 5.793 \\
            
            \emph{State First}
            & 3.724 & \best{0.747} & 13.30 & 5.788 \\
            
            \rowcolor{graybg}
            \emph{Interaction}
            & \best{3.532} & 0.745 & 13.30 & \best{5.779} \\
            \bottomrule
        \end{tabularx}
    \end{minipage}
    \hfill
    \begin{minipage}[t]{0.49\linewidth}
        \centering
        \caption{
            Ablation on the CFG activation interval.
            All variants use Hybrid CFG with \emph{Interaction} composition, setting
            $(w_{\mathrm{cond}}, w_{\mathrm{hist}})=(1.5,1.5)$ inside the interval and $(1,1)$ outside.
            \label{tab:cfg_interval}
        }
        \begin{tabularx}{\linewidth}{@{}l YYYY@{}}
            \toprule
            \textbf{Interval}
            & \textbf{FGD}\,$\downarrow$
            & \textbf{BC}\,$\uparrow$
            & \textbf{Div}\,$\uparrow$
            & \textbf{MSE}\,$\downarrow$ \\
            \midrule
            $[0.0,0.6]$
            & 4.515 & \best{0.751} & 13.59 & 5.782 \\
    
            \rowcolor{graybg}
            $\mathbf{[0.1,0.6]}$
            & \best{3.532} & 0.745 & 13.30 & 5.779 \\
    
            $[0.1,0.7]$
            & 3.706 & 0.746 & 13.60 & 5.786 \\
    
            $[0.1,1.0]$
            & 4.089 & 0.743 & \best{13.75} & \best{5.773} \\
            \bottomrule
        \end{tabularx}
    \end{minipage}
\end{table}

\paragraph{Classifier-Free Guidance Activation Interval.}
%
Following the observation that classifier-free guidance is mainly beneficial in a mid-noise interval~\cite{kynkaanniemi2024applying}, we restrict CFG to this range. 
Concretely, this is implemented by setting the timestep-dependent scales (\eg, $w_{\mathrm{cond}}(t)$) to non-trivial values only within the chosen interval, and to $1$ otherwise (\ie, no guidance).
\Cref{tab:cfg_interval} validates this design by varying the activation interval.
Among the tested cases, $[0.1,0.6]$ achieves the best FGD.
%
Extending the interval toward either the high-noise side (\eg, $[0.1, 0.7]$) or the low-noise side (\eg, $[0.0, 0.6]$) degrades FGD.
We therefore use $[0.1,0.6]$ as the default activation interval.

\begin{table}[t]
    \centering
    \caption{
    \added{Effect of inference-time CFG on the main BEAT2 results.
    \label{tab:cfg_main_results}
    }}

    \renewcommand{\arraystretch}{1.25}
    \setlength{\tabcolsep}{4.5pt}
    \ifmarkadditions\color{blue}\fi
    \arrayrulecolor{black}

    \begin{tabular}{l*{8}{c}}
        \toprule
        \multirow{2}{*}{\textbf{Guidance}}
        & \multicolumn{4}{c}{\textbf{1 Speaker}}
        & \multicolumn{4}{c}{\textbf{All Speakers}} \\
        \cmidrule(lr){2-5}
        \cmidrule(lr){6-9}
        & \textbf{FGD}\,$\downarrow$
        & \textbf{BC}\,$\uparrow$
        & \textbf{Div}\,$\uparrow$
        & \textbf{MSE}\,$\downarrow$
        & \textbf{FGD}\,$\downarrow$
        & \textbf{BC}\,$\uparrow$
        & \textbf{Div}\,$\uparrow$
        & \textbf{MSE}\,$\downarrow$ \\
        \midrule

        w/o CFG
        & 3.76 & 0.744 & \best{13.80} & 5.788
        & 2.59 & 0.639 & \best{9.48} & 3.802 \\

        \rowcolor{graybg}
        Hybrid CFG
        & \best{3.71} & \best{0.746} & 13.38 & \best{5.768}
        & \best{2.33} & \best{0.645} & 9.42 & \best{3.793} \\

        \bottomrule
    \end{tabular}
\end{table}

\paragraph{\added{Effect of CFG on Main Results.}}
\added{\Cref{tab:cfg_main_results} compares PASPA with and without CFG under the $1{,}000$-epoch main-experiment configuration, complementing the $500$-epoch ablation in \cref{tab:ablation_cfg_main}.}
\added{Compared with Hybrid CFG, PASPA without CFG yields worse FGD scores: $3.76$ vs.\ $3.71$ in the 1 Speaker setting and $2.59$ vs.\ $2.33$ in the All Speakers setting.}
\added{Hybrid CFG improves FGD, BC, and MSE in both settings, with slightly lower Div.
Since FGD is our primary quality metric, we use Hybrid CFG.}

\subsection{Additional Analysis of Our Method}

\begin{figure}[t] 
    \centering
    \includegraphics[width=\linewidth]{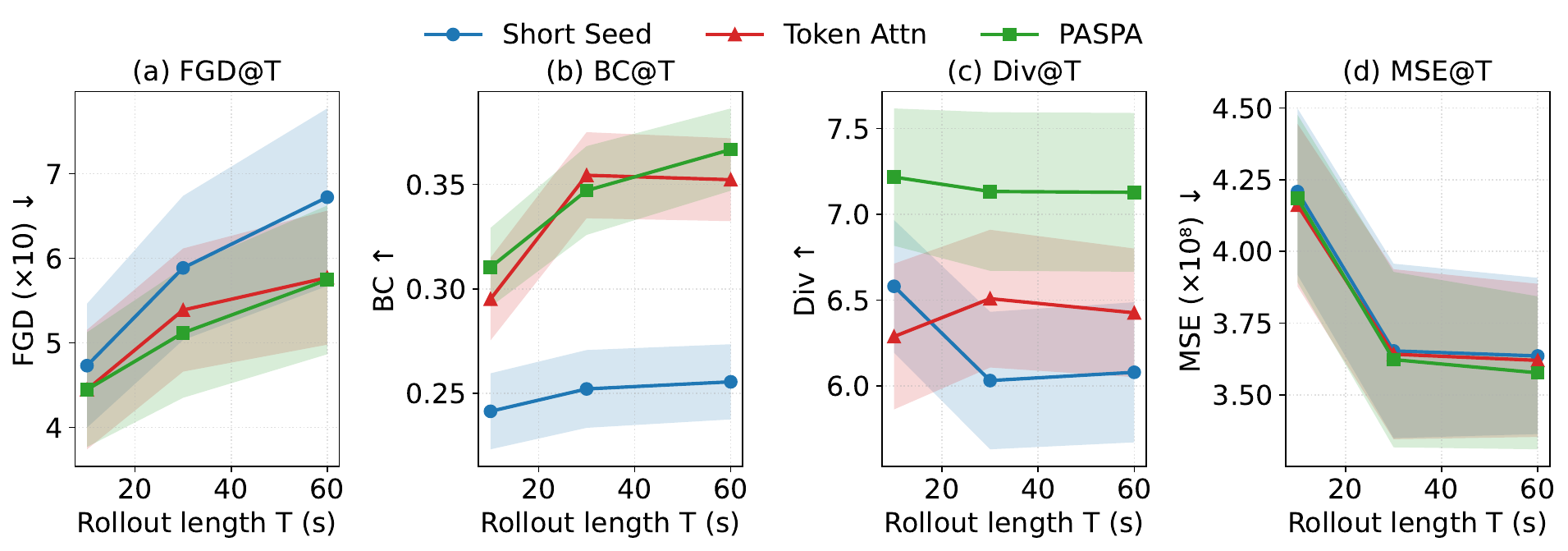}
    \caption{
        Long-rollout endpoint metrics across $T \in \{10, 30, 60\}\,$s, evaluated on the terminal $10$\,s. 
        Shaded regions denote $95\%$ Wald confidence intervals estimated from clip-level bootstrap resampling.
       \label{fig:app_long_rollout_curves}
    }
\end{figure}

\paragraph{Additional Long-Rollout Curves.}
\added{\Cref{fig:app_long_rollout_curves} extends the long-rollout results in \cref{tab:long_rollout}.}
\added{It plots endpoint FGD/Div across rollout horizons and additionally reports BC and MSE.}
%
%
Consistent with \cref{tab:long_rollout}, PASPA shows the weakest FGD degradation and maintains the highest Div throughout the rollout, supporting its advantage in long-horizon generation.
PASPA also achieves competitive BC and slightly lower MSE at later horizons, further supporting its advantage in long-horizon generation.



\begin{table}[t]
    \centering
    \caption{\added{Generation-seed variability in the All Speakers setting. Only the generation seed varies across 50 runs; CIs are $95\%$ Student-$t$ intervals for the mean.}}
    \label{tab:seed_main_results}

    \renewcommand{\arraystretch}{1.20}
    \setlength{\tabcolsep}{8pt}
    \ifmarkadditions\color{blue}\fi
    \arrayrulecolor{black}

    \begin{tabular}{lcccc}
        \toprule
        \multirow{2}{*}{\textbf{Metric}}
        & \multirow{2}{*}{\textbf{Main result}}
        & \multicolumn{3}{c}{\textbf{50 generation seeds}} \\
        \cmidrule(lr){3-5}
        & & \textbf{Mean}
        & \textbf{Sample std.}
        & \textbf{95\% CI of mean} \\
        \midrule

        FGD\,$\downarrow$
        & 2.33 & 2.237 & 0.074 & $[2.216,\,2.258]$ \\

        BC\,$\uparrow$
        & 0.645 & 0.6407 & 0.0015 & $[0.6402,\,0.6411]$ \\

        Div\,$\uparrow$
        & 9.42 & 9.273 & 0.079 & $[9.251,\,9.296]$ \\

        MSE\,$\downarrow$
        & 3.793 & 3.761 & 0.003 & $[3.760,\,3.762]$ \\

        \bottomrule
    \end{tabular}
\end{table}

\paragraph{\added{Generation-Seed Robustness.}}
\added{\Cref{tab:seed_main_results} summarizes the 50-seed All Speakers evaluation under the final configuration.}
\added{Notably, the main-table result in \cref{tab:comparison} comes from the original evaluation, whereas the seed sweep re-evaluates the same checkpoint with freshly initialized generation seeds.}
\added{PASPA achieves a mean FGD of $2.237$ with a standard deviation of $0.074$, indicating limited sensitivity to generation randomness under this fixed-checkpoint setting.}
\added{This consistency supports the reliability of the main FGD result beyond a single favorable sampling run.}

\subsection{User Study}
\label{app:user_study}

\begin{table}[t]
\centering
\caption{
    User study results.
    Each cell reports the mean rank $\pm$ standard deviation across all participants and clips (lower is better; $1$ = best, $4$ = worst).
    %
    \label{tab:user_study_rank}
}
\renewcommand{\arraystretch}{1.20}
\newcolumntype{Y}{>{\centering\arraybackslash}X}
\begin{tabularx}{\linewidth}{lYYYYY}
    \toprule
    Method & Realism & Semantic & Sync & Diversity & Overall \\
    \midrule
    EMAGE      & 2.63$\,\pm\,$1.09 & 2.41$\,\pm\,$1.13 & 2.49$\,\pm\,$1.08 & 2.50$\,\pm\,$1.09 & 2.51$\,\pm\,$1.10 \\
    GestureLSM & 2.60$\,\pm\,$1.15 & 2.80$\,\pm\,$1.12 & 2.82$\,\pm\,$1.09 & 2.83$\,\pm\,$1.12 & 2.76$\,\pm\,$1.12 \\
    SemTalk    & 2.63$\,\pm\,$1.04 & 2.59$\,\pm\,$1.01 & 2.58$\,\pm\,$1.08 & 2.57$\,\pm\,$0.99 & 2.59$\,\pm\,$1.03 \\
    PASPA      & \best{2.14$\,\pm\,$1.12} & \best{2.20$\,\pm\,$1.12} & \best{2.11$\,\pm\,$1.11} & \best{2.10$\,\pm\,$1.14} & \best{2.14$\,\pm\,$1.12} \\
\bottomrule
\end{tabularx}
\end{table}

We conduct a blind ranking study to assess perceptual quality beyond automatic metrics (setup details in \cref{app:imple_user_study}).
\Cref{tab:user_study_rank} reports the mean rank, where PASPA achieves the best on all four criteria, with an overall rank of $2.14$ clearly ahead of the second-best EMAGE ($2.51$).
%
%
\added{These results are consistent with the top-$1$ win rates reported in the main paper (see \cref{fig:user_study_winrate}).}
The user study, together with the metrics in \cref{tab:comparison} and qualitative comparisons in \cref{fig:visual_results}, as well as the video demos in the supplementary material, jointly supports PASPA's advantage over current SOTA methods.

\section{Limitations, Future Work, Failure Cases and Broader Impacts}
\label{app:limitations_future_work}

\subsection{Limitations and Future Work}
\label{app:limitations}

While our method shows promising results in co-speech motion generation, it still has limitations.

\paragraph{Generalization to Other Motion Domains.}
The proposed \emph{past-as-state, present-as-attention} principle is derived from observations on co-speech motion generation, and we have not verified whether the same asymmetry holds in other long-horizon motion domains such as music-to-dance~\cite{li2021ai} or text-to-motion~\cite{tevet2022human}. 
Even if this asymmetry generalizes, the specific PASPA instantiation may require adaptation, and its effectiveness in these broader settings remains to be validated.
We leave the empirical verification of PASPA on other motion generation domains to future work.

\paragraph{Bounded State Capacity.}
PASPA propagates past history through a persistent state of fixed dimensionality $\bfS \in \bbR^{d_v \times d_k}$, regardless of how much history precedes the current block. 
This bounded memory imposes an intrinsic compression limit: for extremely long sequences, older context may be gradually overwritten by new updates.
\added{The state is designed to retain slowly varying style statistics (\eg, gesture amplitude and spatial preference), but may lose details of distant gestures or long-range correspondences needed to reproduce earlier motions.}
This trade-off is fundamental to the broader family of linear recurrent models~\cite{yang2024gated,yang2025gated,dao2024transformers,gu2023mamba}. 
A promising direction is the recently proposed log-linear attention~\cite{guo2026loglinear}, which replaces the fixed-size state with a logarithmically growing set of states and directly applies to Gated DeltaNet. 
\added{We leave this for future work.}

\paragraph{Train-Inference Gap.}
Like most autoregressive generation methods, PASPA is trained with teacher forcing method (each block is generated conditioned on ground-truth history)~\cite{ning2024elucidating,schmidt2019generalization}, while inference must rely on the previously generated blocks, leading to exposure bias. 
This bias may lead to error accumulation in long rollouts, \added{potentially manifesting as gesture repetition, posture collapse, identity drift, or global-translation artifacts.} 
We do not explore recent techniques that explicitly bridge this gap, such as Self Forcing~\cite{huang2025self}, which performs autoregressive rollout during training to align the training and inference distributions. 
Integrating such strategies is a natural direction for future work.

\paragraph{\added{Evaluation Limitations.}}
\added{Our quantitative evaluation uses automatic metrics, yet no universally accepted metric fully captures long-term stylistic consistency and perceptual quality.}
\added{We therefore complement these metrics with a user study to assess PASPA's effectiveness.}
\added{Developing new style or perceptual metrics is beyond our scope and is left for future work.}
\added{Moreover, the generalizability of these findings should be further validated across broader real-world deployment scenarios.}

\subsection{Failure Cases}
\label{app:failure_cases}

\begin{figure}[t] 
    \centering
    \includegraphics[width=\linewidth]{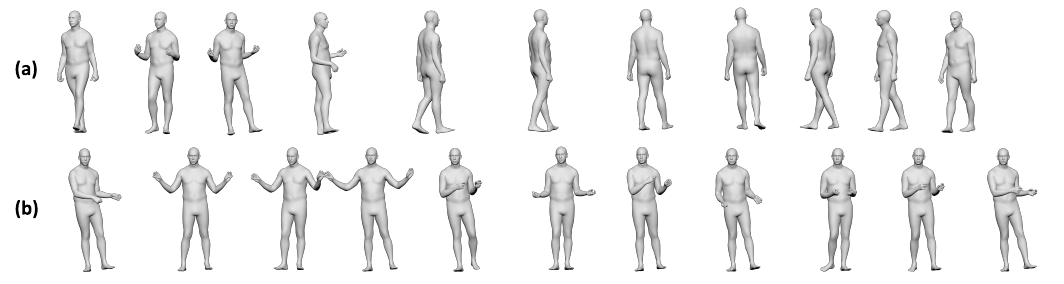}
    \caption{
        Failure case on sequence \texttt{16\_jorge\_1\_3\_3}. Time progresses 
        from left to right. (a) Ground-truth. (b) Our generated sequence.   
        \label{fig:fail}
    }
\end{figure}

\Cref{fig:fail} shows a representative failure mode of our method.
In the ground truth, the speaker performs a large in-place circling motion, leading to strong and highly non-linear changes in global translation.
Although PASPA generates plausible local gestures, it fails to recover this circling trajectory and produces motion that remains mostly around a fixed spatial region.
We attribute this limitation to the current motion generation pipeline~\cite{zhang2026mitigating}, where global translation is learned together with joint rotations (\ie, rot6d~\cite{zhou2019continuity}) in the same motion stream.
These two components have different geometric structures: translation lies in the Euclidean space $\mathbb{R}^{3}$, whereas the 6D rotation representation (\ie, rot6d~\cite{zhou2019continuity}) parameterizes rotations through constrained orthonormal frames on the Stiefel manifold, corresponding to the structure of $\mathrm{SO}(3)$.
Learning them jointly, especially within the torso branch, may underemphasize the distinct dynamics of large-amplitude global translation.
Following GlobalDiff's co-speech motion generation architecture~\cite{zhang2026mitigating}, we keep it for consistency, but more accurate modeling of global translation remains an important direction for future work.

\subsection{Broader Impacts}
\label{app:broader_impacts}

\paragraph{Positive Impacts.}
PASPA aims to improve long-horizon co-speech motion generation, which can benefit digital humans~\cite{zhu2023human}, virtual assistants~\cite{yoon2020speech}, animated characters~\cite{habibie2021learning}, education, and entertainment by enabling more coherent and expressive nonverbal behavior.
By reducing the manual effort required for full-body animation, it may also lower the production cost of educational content, virtual avatars, and interactive media.

\paragraph{Negative Impacts.}
Like other generative models for human motion and appearance~\cite{tevet2022human,liu2024emage,yi2023generating}, PASPA could be misused to create misleading speech-driven avatars, contributing to impersonation, disinformation, or non-consensual digital likenesses.
Improved long-horizon coherence and style preservation may make synthetic embodied behavior harder to distinguish from real recordings.
Although our method generates motion rather than audio or appearance, this risk remains relevant when motion is combined with separately synthesized voice, face, or identity generation systems.

\paragraph{Mitigation.}
Our experiments are conducted in a research setting using public co-speech motion datasets adopted in prior work~\cite{liu2024emage}, and our contribution focuses on improving motion generation within this established benchmark setting, rather than identity recognition, surveillance, or high-stakes decision making.
We do not collect new identifiable personal data or deploy the model in user-facing applications.
When releasing code or checkpoints, we will use research-oriented license terms and encourage users to avoid deceptive, non-consensual, or identity-misrepresenting applications.

\section{Licenses of Used Assets}
\label{sec:licenses_used_assets}
\paragraph{Datasets.} We summarize the \added{datasets} used in this work and \added{their licenses} below. 
\begin{itemize}
    \item \textbf{BEAT2}~\cite{liu2024emage}: This dataset is released under the Apache 2.0 license.
    \item \added{\textbf{SHOW}~\cite{yi2023generating}:
    Provided under the project's {Data \& Software Copyright License for non-commercial scientific research purposes}.}
\end{itemize}

\paragraph{Code and Models.}
We build upon the following public codebases and pretrained models.
\begin{itemize}
    \item \textbf{SANA}~\cite{xie2024sana}: Released under the Apache 2.0 license. We adopt its codebase as the backbone for building our flow-matching framework.
    
    \item \textbf{GlobalDiff}~\cite{zhang2026mitigating}: 
    Released under the Apache 2.0 license. 
    We use its pretrained audio-to-expression weights \added{for BEAT2}, as well as parts of its data loading pipeline, global-rotation forward-kinematics logic, loss functions, and network architecture.
    
    \item \textbf{Flash Linear Attention (FLA)}~\cite{yang2024fla}: 
    Released under the MIT license. 
    We adopt its Triton-based Gated DeltaNet (GDN)~\cite{yang2025gated} kernel and PyTorch-based GDN layer implementation, and extend the Triton kernel to support our block-causal variant.
    
    \item \textbf{Self-Forcing}~\cite{huang2025self}: 
    Released under the Apache 2.0 license. 
    We adapt its autoregressive-diffusion inference pipeline to our setting.
    
    \item \textbf{SMPL-X}~\cite{pavlakos2019expressive}: 
    Released under the SMPL-X non-commercial scientific research license. 
    We use the SMPL-X body model and its Blender add-on for constructing SMPL-X-compatible digital human representations.
    
    
    \item \textbf{Emage Evaltools}~\cite{liu2024emage}: 
    Released under the Apache 2.0 license. 
    We use it for computing evaluation metrics.
    
    \item \added{\textbf{TalkSHOW}~\cite{yi2023generating}:
    We adapt its face-prediction implementation to WavLM features and train the predictor on SHOW. No repository-level license file is provided; the project website lists non-commercial research terms.}
    
    \item \added{\textbf{DiffSHEG}~\cite{chen2024diffsheg}:
    Its code is released under the BSD 3-Clause license. We use its public evaluation components and pretrained feature extractors for SHOW evaluation.}
    
    \item \added{\textbf{WavLM}~\cite{chen2022wavlm}:
    Its official implementation is released under the MIT license. We use a frozen WavLM-Large encoder to extract speech features for BEAT2 and SHOW.}
\end{itemize}

All assets are properly cited in the main text and used in compliance with their respective licenses for academic research. 

\newpage
\section*{NeurIPS Paper Checklist}

\begin{enumerate}

\item {\bf Claims}
    \item[] Question: Do the main claims made in the abstract and introduction accurately reflect the paper's contributions and scope?
    \item[] Answer: \answerYes{} 
    \item[] Justification: 
    The abstract and introduction in \cref{sec:introduction} clearly state our scope and contributions for long-horizon co-speech motion generation:
    (1) identifying the past-present asymmetry and formulating the \emph{past-as-state, present-as-attention} principle;
    (2) proposing the \textbf{PASPA} module with inter-block state modeling and intra-block attention;
    (3) developing a blockwise flow matching framework with tailored training and inference.
    %
    These claims are developed in \cref{sec:methodology}, supported by main results in \cref{sec:comparison}, and validated by ablations in \cref{sec:ablation}.

\item {\bf Limitations}
    \item[] Question: Does the paper discuss the limitations of the work performed by the authors?
    \item[] Answer: \answerYes{}
    \item[] Justification: 
    We discuss limitations in \cref{app:limitations_future_work}, including the unverified generalization of the past-as-state, present-as-attention principle to other motion domains, the bounded capacity of the fixed-size state, and the teacher-forcing train-inference gap that may cause exposure bias in long rollouts.
    We also provide future works and failure case analysis in \cref{app:limitations,app:failure_cases}, respectively.

\item {\bf Theory assumptions and proofs}
    \item[] Question: For each theoretical result, does the paper provide the full set of assumptions and a complete (and correct) proof?
    \item[] Answer: \answerNA{}
    \item[] Justification: 
    The paper does not present formal theoretical results such as theorems or lemmas requiring proof.
    The mathematical expressions are used to define the method and analysis, including the FM path in \cref{eq:flow_matching_inter}, the BGDN update/read rules in \cref{eq:gdn_write,eq:gdn_and_bgdn_read}, the BMHSA block mask in \cref{eq:bmhsa_attn}, the blockwise FM formulation in \cref{eq:bfm_path,eq:bfm_network}, the training objectives in \cref{eq:bfm_total_loss,eq:bfm_bd_loss}.

\item {\bf Experimental result reproducibility}
    \item[] Question: Does the paper fully disclose all the information needed to reproduce the main experimental results of the paper to the extent that it affects the main claims and/or conclusions of the paper (regardless of whether the code and data are provided or not)?
    \item[] Answer: \answerYes{} 
    \item[] Justification: 
    We describe the core method in sufficient detail for reproduction: the architecture is specified in \cref{fig:network,sec:architeture}; the blockwise FM formulation is given in \cref{sec:blockwise_flow_matching}; the loss objectives are defined in \cref{eq:bfm_total_loss,eq:bfm_bd_loss} with full details in \cref{app:loss_details}; and the long-horizon sampling procedure is specified by \cref{alg:rollout_cfg} and \cref{sec:inference}.
    %
    \added{We also report the experimental setup, including metrics, in \cref{sec:experiments}, the datasets in \cref{app:dataset_information}, and the full training/inference configuration in \cref{app:implementation_details}.
    \Cref{tab:training_details} summarizes the BEAT2 configurations, with SHOW-specific adaptations described in \cref{app:imple_main_results}.}

\item {\bf Open access to data and code}
    \item[] Question: Does the paper provide open access to the data and code, with sufficient instructions to faithfully reproduce the main experimental results, as described in supplemental material?
    \item[] Answer:  \answerYes{}
    \item[] Justification: 
    We will release the code with instructions for environment setup, data preparation, training, inference, and evaluation upon publication.
    %
    \added{The experiments use BEAT2 and SHOW, obtained through their respective official releases and access procedures. 
    Dataset descriptions and representation conversions are provided in \cref{app:dataset_information,app:imple_main_results}.}

\item {\bf Experimental setting/details}
    \item[] Question: Does the paper specify all the training and test details (e.g., data splits, hyperparameters, how they were chosen, type of optimizer) necessary to understand the results?
    \item[] Answer: \answerYes{}
    \item[] Justification: 
    The core paper specifies the main experimental setting in \cref{sec:experiments}, including the train/test splits, evaluation subsets, optimizers and evaluation metrics.
    %
    %
    \added{Implementation details for both datasets, including shared settings and dataset-specific adaptations, are provided in \cref{app:implementation_details}. \Cref{tab:training_details} summarizes the BEAT2 configurations.}
    Upon code release, we will provide the configuration files and scripts used for the reported experiments, including YAML and shell-script based training and evaluation settings.

\item {\bf Experiment statistical significance}
    \item[] Question: Does the paper report error bars suitably and correctly defined or other appropriate information about the statistical significance of the experiments?
    \item[] Answer: \answerYes{}. 
    \item[] Justification: 
     We report uncertainty or variability for key analyses. \Cref{fig:acf} uses 95\% bootstrap confidence intervals for autocorrelation estimates, \cref{fig:app_long_rollout_curves} uses 95\% confidence intervals from clip-level bootstrap resampling for long-rollout metrics, and \cref{fig:infer_efficiency} reports median values with min-max latency bands over repeated runs. 
     For the user study, \cref{tab:user_study_rank} reports mean rank with standard deviation across valid ranking groups.
     \added{\Cref{tab:seed_main_results} reports means, standard deviations, and $95\%$ Student-$t$ confidence intervals across $50$ generation seeds.}
    %
     Details are provided in the corresponding captions and \cref{app:imple_other_exp,app:imple_user_study}.

\item {\bf Experiments compute resources}
    \item[] Question: For each experiment, does the paper provide sufficient information on the computer resources (type of compute workers, memory, time of execution) needed to reproduce the experiments?
    \item[] Answer: 
    \answerYes{}.
    \item[] Justification: 
    We report the compute resources in \cref{tab:training_details} and \cref{app:implementation_details}.
    The \added{BEAT2} model is trained on 4 NVIDIA A800 GPUs (40GB) using Float32 precision for 1000 epochs, taking approximately 30 hours (\ie, 120 GPU-hours).
    \added{The SHOW motion model is trained for 1000 epochs (82{,}000 steps), using the same batch-size and precision settings.}
    We also report the model size, MACs, batch size, and other optimization settings in \cref{app:implementation_details}. 
    %
    %
    \added{Models for the controlled ablations are trained for $500$ epochs using the same hardware and configuration.}
    \added{The generation-seed evaluation in \cref{tab:seed_main_results} reuses the final model without additional training.}
    %
    For the inference-efficiency benchmark, we specify the single-GPU setup and experimental protocol in \cref{app:imple_other_exp}.
    
\item {\bf Code of ethics}
    \item[] Question: Does the research conducted in the paper conform, in every respect, with the NeurIPS Code of Ethics \url{https://neurips.cc/public/EthicsGuidelines}?
    \item[] Answer: \answerYes{}
    \item[] Justification: 
    We have reviewed the NeurIPS Code of Ethics and believe that the research conforms to it.
    Our experiments use \added{public co-speech motion benchmarks} under their intended research setting, and the work focuses on motion generation methodology rather than identity recognition, surveillance, or high-stakes decision making.

\item {\bf Broader impacts}
    \item[] Question: Does the paper discuss both potential positive societal impacts and negative societal impacts of the work performed?
    \item[] Answer: \answerYes{}
    \item[] Justification:
    We discuss broader impacts in \cref{app:broader_impacts}, including positive applications in domains such as digital humans, virtual assistants, and animated characters, as well as reduced animation production cost. %
    We also discuss potential negative impacts such as misleading speech-driven avatars, impersonation, disinformation, and non-consensual digital likenesses, and clarify that our experiments are conducted in a research setting using public datasets without collecting new identifiable personal data or deploying user-facing systems.
    
\item {\bf Safeguards}
    \item[] Question: Does the paper describe safeguards that have been put in place for responsible release of data or models that have a high risk for misuse (e.g., pre-trained language models, image generators, or scraped datasets)?
    \item[] Answer: \answerNA{}
    \item[] Justification: 
    The paper does not release a new scraped dataset or a high-risk generative model such as a language, image, voice, or identity generator.
    Our method focuses on motion generation for research purposes, and does not generate language, image, video, audio, photorealistic appearance, or identity information.
    Broader misuse considerations are discussed in \cref{app:broader_impacts}.

\item {\bf Licenses for existing assets}
    \item[] Question: Are the creators or original owners of assets (e.g., code, data, models), used in the paper, properly credited and are the license and terms of use explicitly mentioned and properly respected?
    \item[] Answer: 
    \answerYes{}
    \item[] Justification: 
    We use publicly available datasets and codebases, and properly cite their original papers in the main text. 
    The complete list of used assets, including their sources and licenses (\eg, Apache 2.0, MIT, BSD 3-Clause, and the SMPL-X non-commercial research license), is summarized in \cref{sec:licenses_used_assets}, and all assets are used in compliance with their respective licenses for academic research.

\item {\bf New assets}
    \item[] Question: Are new assets introduced in the paper well documented and is the documentation provided alongside the assets?
    \item[] Answer: \answerYes{}
    \item[] Justification:
    \added{We release the PASPA code, trained model checkpoints, and generated motion results, with accompanying documentation covering environment setup, data preparation, training, inference, and evaluation, as well as licensing information and limitations.}

\item {\bf Crowdsourcing and research with human subjects}
    \item[] Question: For crowdsourcing experiments and research with human subjects, does the paper include the full text of instructions given to participants and screenshots, if applicable, as well as details about compensation (if any)? 
    \item[] Answer: \answerYes{}
    \item[] Justification:
    The paper includes a user study for perceptual evaluation in \cref{app:user_study}. 
    We provide full participant instructions in \cref{app:user_study} and a screenshot of the evaluation interface in \cref{fig:user_study_page}.
    Participants were volunteers recruited within our institution, without monetary compensation.
    The user study follows the NeurIPS Code of Ethics.

\item {\bf Institutional review board (IRB) approvals or equivalent for research with human subjects}
    \item[] Question: Does the paper describe potential risks incurred by study participants, whether such risks were disclosed to the subjects, and whether Institutional Review Board (IRB) approvals (or an equivalent approval/review based on the requirements of your country or institution) were obtained?
    \item[] Answer: 
    \answerYes{}
    \item[] Justification:
    The user study involves only voluntary ranking of short generated motion clips, with no personal data collected and no risk to participants beyond ordinary computer use. 
    Participants were volunteers who were informed of the study purpose and voluntary nature of participation. 
    Under our institution's review process, this type of low-risk perceptual study with informed consent does not require formal IRB approval. 
    The study follows the NeurIPS Code of Ethics.

\item {\bf Declaration of LLM usage}
    \item[] Question: Does the paper describe the usage of LLMs if it is an important, original, or non-standard component of the core methods in this research? Note that if the LLM is used only for writing, editing, or formatting purposes and does \emph{not} impact the core methodology, scientific rigor, or originality of the research, declaration is not required.
    \item[] Answer: \answerNA{}
    \item[] Justification: 
    The core method development and experiments in this paper do not use LLMs as important, original, or non-standard components.
    Any use of LLMs, if any, was limited to writing, editing, or formatting assistance and did not affect the methodology, experiments, scientific claims, or originality of the work.

\end{enumerate}

\end{document}